\documentclass[conf]{new-aiaa}
\usepackage[utf8]{inputenc}
\usepackage{float}
\usepackage{graphicx}
\usepackage{amsmath}
\usepackage[version=4]{mhchem}
\usepackage{siunitx}
\usepackage{placeins}
\usepackage{booktabs}
\usepackage{longtable,tabularx}
\usepackage{url}
\usepackage{xcolor}
\definecolor{opengreen}{RGB}{0,120,60}
\definecolor{closedred}{RGB}{178,34,34}

\usepackage{subcaption}
\usepackage{threeparttable}
\usepackage{wrapfig}
\usepackage[most]{tcolorbox}
\usepackage{caption}
\usepackage{hyperref}
\usepackage{float}
\usepackage[ruled,vlined]{algorithm2e}
\usepackage{xcolor}
\definecolor{pilotbadge}{HTML}{2A6FB0}
\definecolor{atcbadge}{HTML}{C0603A}
\definecolor{pilotbg}{RGB}{230,245,255}
\definecolor{atcbg}{RGB}{255,238,220}
\usepackage{hyperref}
\hypersetup{
    colorlinks=true,
    urlcolor=blue,
    linkcolor=blue,
    citecolor=blue
}
\usepackage{booktabs}
\usepackage{tabularx}
\usepackage{array}
\usepackage{fontawesome5}
\usepackage[most]{tcolorbox}

\usepackage{booktabs}
\usepackage{siunitx}
\usepackage[table]{xcolor}
\usepackage{array}

\definecolor{pilotblue}{RGB}{120,185,225}
\definecolor{atcpink}{RGB}{220,150,180}
\newtcolorbox{pilotbox}{
  colback=pilotblue,
  colframe=pilotblue,
  coltext=white,
  arc=3mm,
  boxrule=0pt,
  left=3mm,right=3mm,top=2mm,bottom=2mm,
  width=\linewidth
}

\newtcolorbox{atcbox}{
  colback=atcpink,
  colframe=atcpink,
  coltext=white,
  arc=3mm,
  boxrule=0pt,
  left=3mm,right=3mm,top=2mm,bottom=2mm,
  width=\linewidth
}

\title{Air Traffic Control Using Large Language Models: Prompt Engineering, Architecture, and Evaluation}

\author{Mahyar Ghazanfari\footnote{Ph.D. Student, Department of Mechanical and Aerospace Engineering, The George Washington University, AIAA Student Member.}}
\affil{George Washington University, Washington, DC 20052, USA}

\author{Matthias Casanova\footnote{Undergraduate Student, Department of Mechanical and Civil Engineering, California Institute of Technology, AIAA Student Member.} and Jordan Kam\footnote{Ph.D. Student, Department of Aerospace Engineering, California Institute of Technology, AIAA Student Member.}}
\affil{California Institute of Technology, Pasadena, CA 91125, USA}

\author{Alex Zongo\footnote{Ph.D. Student, Department of Mechanical and Aerospace Engineering, The George Washington University, AIAA Student Member.} and Peng Wei\footnote{Professor, Department of Mechanical and Aerospace Engineering, The George Washington University, AIAA Associate Fellow.}}
\affil{George Washington University, Washington, DC 20052, USA}

\author{Torsten Darrell\footnote{Research Intern, Bayen Lab, University of California, Berkeley, AIAA Student Member. } and Alexandre Bayen\footnote{Professor, Department of Electrical Engineering and Computer Sciences, University of California, Berkeley.}}
\affil{University of California, Berkeley, CA 94720, USA}

\begin{document}

\maketitle

\textbf{Air traffic control (ATC) communication is a safety-critical dialogue that
remains largely human-driven even as other parts of air traffic management have been semi-automated. In this article, we experimentally evaluate whether large language models (LLMs) can generate
operationally realistic ATC transmissions. An experimental general-aviation flight
flying over the San Francisco ``Bay Tour'' route is hand-transcribed and used as ground truth
(P0). Through a pilot-in-the-loop process we design five prompt structures (P1--P5)
of increasing constraint and embed them in a stateful multi-turn pipeline, where the
model plays ATC to a fixed pilot transcript while conditioning on the
accumulating dialogue history. Across nine open- and closed-source LLMs we vary the
prompt, the presence of a worked transcript from a different experimental flight as an
in-context example, and whether the model conditions on its own prior replies or on injected ground-truth history. Turns are scored with lexical,
structural, and semantic similarity metrics and by an LLM-as-judge (GPT-5.5)
validated against human expert annotation. Supplying a worked example improves similarity, but tightening the prompt does not: the lightest prompts perform best and the most heavily scripted one collapses as its
own errors accumulate through the dialogue, which injecting correct history repairs.
These results outline a concrete
path and its current limits toward LLM-assisted ATC. Code and data can be found \href{https://github.com/AeroMatt5/ATC_LLM_Prompting}{here}.}

\section{Introduction}

Air traffic control (ATC) is one of the most demanding and safety-critical jobs in the US
transportation network \cite{jou2013study}. Human-to-human communication is the backbone of this system and provides the standardized format that controllers and pilots use to keep aircraft separated, manage the flow of traffic, and coordinate
operations across busy airspace \cite{icao4444, faa7110}. Unlike everyday conversation,
ATC speech follows a tightly constrained phraseology, set by the Federal Aviation
Administration (FAA), that is meant to remove ambiguity, reduce cognitive load, and
support fast decisions under pressure. As air traffic continues to grow, ATC must
handle denser traffic while holding strict safety margins. Yet ATC communication remains
one of the most human-intensive parts of the air traffic management (ATM) system, built
on voice radio, situational awareness, and procedural discipline rather than automation
\cite{atco2corpus}. This reliance on human communication is both a strength and a limit
on how far current operations can scale in the National Airspace.

These pressures are becoming more urgent with the rise of Advanced Air Mobility (AAM)
\cite{dot_aam_strategy_2025}: urban air mobility, dense low-altitude operations, electric
vertical takeoff and landing (eVTOL) aircraft, uncrewed systems, and other new airspace
users \cite{kopardekar2016uam, thipphavong2018uam, sengupta2025uam}. Today's ATC
architecture was not built to scale to the volume and diversity of operations that these
concepts imply. As a result, researchers are exploring next-generation ATM designs that
combine automation, distributed decision-making, and digital communication while keeping
safety intact. In this setting, AI-driven communication tools may help human operators
coordinate with a growing number of increasingly autonomous aircraft. Understanding how
well foundation models handle realistic, multi-turn ATC dialogue is therefore an
important first step.

This article takes that step. We measure how closely large language models (LLMs)
reproduce human controller behavior in an operational dialogue, and we ask what it takes
to evaluate such a system credibly. Our contributions are:

\begin{itemize}
  \item \textbf{A ground-truth benchmark and stateful evaluation pipeline.} We
  hand-transcribe an experimental general-aviation flight flying the San Francisco ``Bay Tour''
  and use it as ground truth (P0). Instead of scoring isolated turns, we run each model
  through the whole conversation, conditioning every reply on the growing dialogue
  history, as a deployed controller would be.
  \item \textbf{A systematic study of prompting and conversational grounding.} We design
  five prompt structures (P1--P5), from a minimal role assignment to heavily constrained
  rule sets, with the strongest prompts refined through pilot feedback. We separate
  prompt quality from conversational grounding by comparing histories built from the
  model's own replies against histories seeded with the true controller transmissions,
  and we measure the effect of adding a complete worked transcript from a different experimental
  flight as an in-context example.
  \item \textbf{A validated evaluation methodology.} We combine standard text-similarity
  metrics with an LLM-as-judge that scores seven operational dimensions, and we validate
  the judge against expert human annotation. This shows which cheap automatic metrics can
  stand in for expert judgment and which cannot.
\end{itemize}

Across nine proprietary and open-source LLMs, we find that a worked in-context example
consistently improves alignment with human ATC, while adding prompt constraints does not.
Beyond a light role-and-context prompt, extra instructions tend to hurt. We also find
that accurate dialogue history is what keeps performance from degrading over a long
conversation, and that it rescues even the most heavily scripted prompt. Among our findings, we demonstrate that current
models reliably reproduce the \emph{form} of controller speech but fall short on its
operational \emph{content}, the distinction that matters most for safety-critical use.

\section{Related Work}

Automation has reshaped many parts of ATM, including surveillance, trajectory prediction,
decision-support tools, and large-scale traffic flow optimization \cite{bayen2006adjoint}.
Automating the controller--pilot conversation, by contrast, has lagged behind, because it
demands strict safety, contextual reasoning, and precise standardized language. Recent
advances in artificial intelligence, and LLMs in particular, open new possibilities by
generating context-aware, human-like responses \cite{chen2023llm}. Unlike older
rule-based systems, LLMs interpret and produce natural language flexibly, which could
support controller assistance, training, and augmentation of existing workflows. At the
same time, putting AI into ATC communication raises hard questions of reliability,
consistency, procedural compliance, and verification, which is why systematic evaluation
against real operational behavior is needed. Within ATM more broadly, automation has
focused on optimization and system-level scalability, including foundational traffic-flow
optimization work \cite{bayen2006adjoint}. These efforts
improved efficiency and predictability at the system level but left controller--pilot
communication a largely human-driven process.

A growing body of work applies natural language processing (NLP) to structure aviation's
traditionally voice- and text-based communication. NASA researchers have used NLP to
classify and interpret Notices to Airmen (NOTAMs), turning dense procedural notices into
actionable information \cite{szeto2024notam}, while the ATCO2 corpus enables large-scale
automatic speech recognition and named-entity extraction, such as callsign, command, and
value, over recorded controller--pilot radio exchanges \cite{atco2corpus}. These efforts
use transformer-based \cite{vaswani2017attention} or domain-adapted models to classify,
structure, or extract information rather than to generate communication. More recently,
domain-adapted foundation models have appeared. AviationGPT continues pre-training open
LLaMA-2 and Mistral backbones on curated aviation text to answer questions and summarize
National Airspace System documents \cite{wang2024aviationgpt}, showing clear momentum
toward aviation-specific language models. Generative applications have followed, most
notably CHATATC, which trains a conversational agent on more than 80{,}000 historical
Ground Delay Program records from 2000--2023 and studies its behavior in a deliberately
\emph{non}-safety-critical, strategic traffic-flow setting \cite{abdulhak2024chatatc}.
Closer to the tactical loop, \cite{andriuskevicius2024embodied}  embody a language model as an
air traffic agent with function-calling and learning, resolving conflicts without human
intervention.

A parallel line of work applies LLMs to aviation planning and safety assessment, using
natural language as the interface between operators and an automated reasoner. \cite{tabrizian2025cot} use chain-of-thought prompting to produce end-to-end flight routes under wind
hazards while eliciting operator preferences in natural language, keeping a human in the
loop. FRAMe extends this by pairing a planner LLM with a
retrieval-augmented memory and a multi-modal coach agent, so that generated plans both
satisfy mission constraints and match operator preferences, reaching up to 93.8\%
aggregate validity across four backbone LLMs \cite{tabrizian2026frame}. Closest to our
study, \cite{darrell2026nontowered} assess safety around non-towered airports by having 
vision-language models (VLMs) reason jointly over transcribed Common Traffic Advisory Frequency (CTAF) calls,
METAR weather, ADS-B trajectories, and VFR sectional charts, benchmarking three open and
three closed models against a twelve-category hazard taxonomy and exceeding a macro
$F_1$ of 0.85 on a nominal-versus-danger task. Together these
works show that LLMs can reason over diverse aviation data and produce operator-aligned
plans. Critically, though, they either analyze communication after the fact or generate
plans rather than transmissions, and none puts the model in the controller's seat over a
live, multi-turn exchange.

The present work targets that gap. We evaluate LLM behavior in a controlled multi-turn ATC
scenario driven by real pilot transmissions from an operational flight recording,
generating repeated stochastic controller responses and comparing them against
ground-truth ATC dialogue across several similarity metrics. Three features set this
evaluation apart. First, it is \emph{stateful}: the model conditions on the growing
dialogue history rather than isolated turns, which lets us separate per-turn competence
from error accumulation over a conversation. Second, it is \emph{prompt-conditioned}: we
compare five systematically varied prompt structures against the same ground truth. Third,
it is \emph{validated}: we pair automatic metrics with an LLM-as-judge rubric and check
that judge against human expert annotation, so reported quality reflects operational
correctness rather than surface overlap. To our knowledge, no prior work has run a
multi-turn, prompt-conditioned, ground-truth-aligned evaluation of LLM-generated ATC
transmissions with human-validated scoring.

\section{Methodology}
\label{sec:method}

\begin{figure}[!t]
  \centering
  \includegraphics[width=\linewidth]{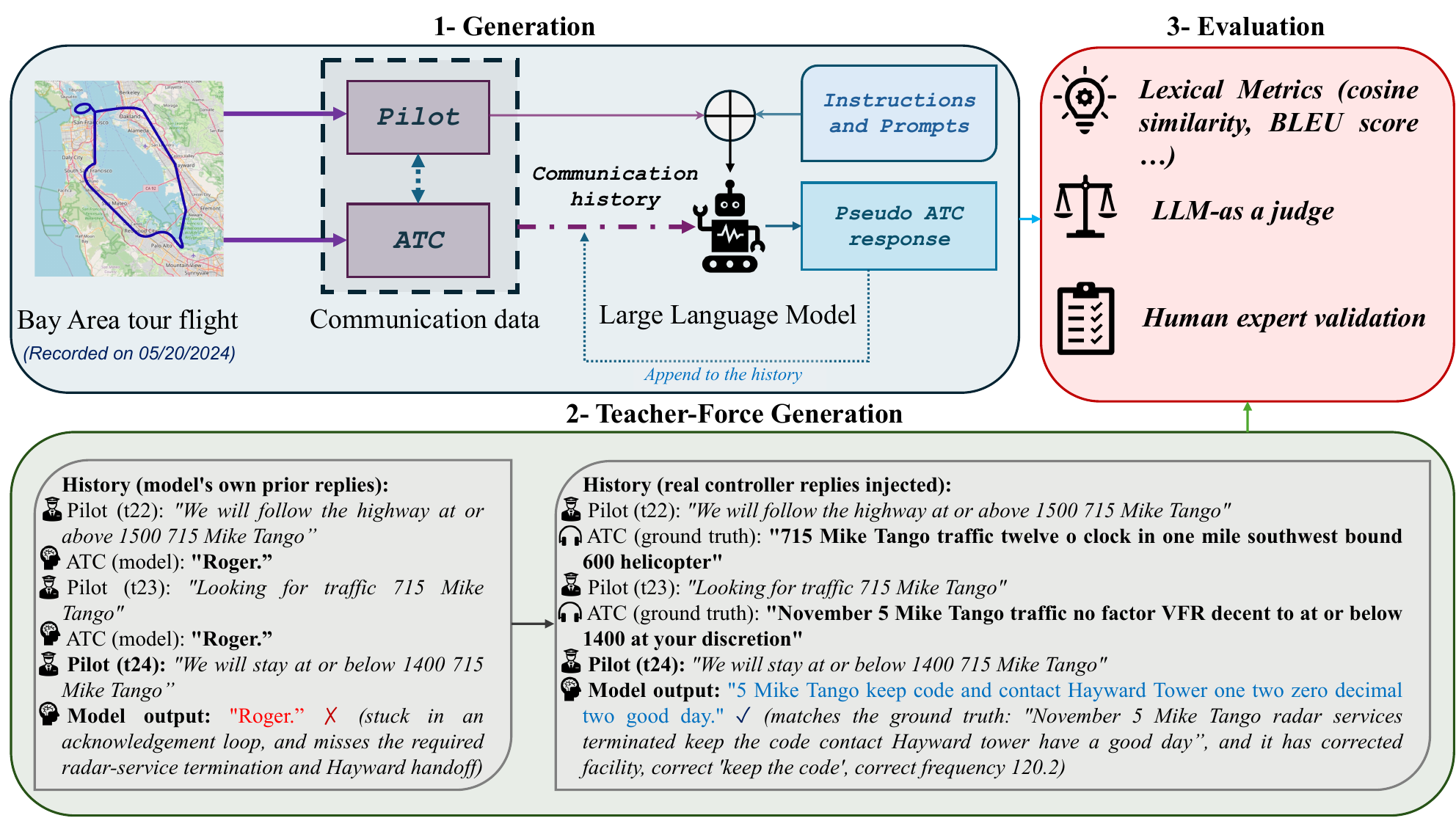}
  \caption{Overall architecture. \textbf{(1)~Generation:} pilot transmissions
  transcribed from an experimental Bay Area tour flight are issued one turn at a time to the
  language model, which is conditioned on a system prompt and on the communication
  history accumulated so far; each pseudo-ATC response is appended back into that
  history. \textbf{(2)~Teacher-forced generation:} the same scenario is replayed with
  the controller side of the history replaced by the ground truth transmissions. The example
  contrasts the two regimes at turn~24. Conditioned on its own replies (left), the
  model has fallen into an Acknowledgment loop and answers ``Roger,'' missing the
  required radar-service termination and Hayward handoff. Given the true history
  (right), the same turn is answered with the correct facility, the correct
  ``keep the code'' instruction, and the correct frequency 120.2.
  \textbf{(3)~Evaluation:} every turn is scored by automatic similarity metrics, by an
  LLM-as-judge, and---on a stratified subsample---by a human expert.}
  \label{fig:architecture}
\end{figure}

Our study is organized as a three-stage pipeline, shown in
Fig.~\ref{fig:architecture}. In the \emph{generation} stage, pilot transmissions
taken verbatim from an experimental general-aviation (GA) flight are fed one at a time to a
language model that has been instructed to act as the air traffic controller; each
model reply is appended to a running dialogue history so that later turns are
answered in context. In the \emph{teacher-forced generation} stage, we repeat the
identical procedure but overwrite the controller side of that history with the ground truth
controller transmissions, which lets us separate a model's ability to answer a single
turn from its ability to survive its own accumulated mistakes. In the
\emph{evaluation} stage, every generated turn is scored three ways: by automatic
similarity metrics against the true controller reply, by an LLM-as-judge applying an
operational rubric, and by a human expert who re-scores a stratified subsample to
validate the judge.

\subsection{Reference Data}
\label{sec:method-data}
Two experimental GA flights are used, and they play strictly separate roles. Complete flight data can be found in Appendix \ref{app:transcripts}.

\paragraph{Evaluation scenario (P0).} We hand-transcribe the complete controller--pilot
exchange of a ``San Francisco Bay Area Flight Tour'' flown by a GA pilot
\cite{youtube_ejUM1YYU0Fc}. The flight is a Cessna (callsign 715 Mike Tango) departing
Palo Alto (KPAO), transitioning the peninsula near San Carlos and the Bay, and
recovering through the Hayward and Oakland area, so that the scenario exercises ground,
tower, and approach interactions as well as facility handoffs. The transcript yields
36 aligned turns, each a (pilot transmission, controller reply) pair indexed
by a turn identifier. This transcript, denoted P0, is the sole ground truth against
which all generated transmissions are scored. It is never shown to a model in any
condition.

\paragraph{In-context example flight.} A \emph{second, different} flight, a
Cirrus (callsign 485 Echo Mike Romeo) operating out of Reid--Hillview, is transcribed
in the same format, yielding 49 turns. Unlike publicly archived ATC audio, this
exchange was captured first-hand: the flight was flown by a co-author of this work,
who recorded the onboard radio audio and the accompanying flight logs in the place, and the
transcript was produced by hand from that recording rather than retrieved from an
online archive. This transcript is used exclusively as a worked in-context
example (Sec.~\ref{sec:method-generation}). Keeping the demonstration flight disjoint
from the evaluation flight is what makes the in-context condition a test of
transferable phraseology rather than of memorization. No model ever sees any part of
P0 as an example.

\subsection{Prompt Design}
\label{sec:method-prompts}
We construct five system prompts, P1--P5, that increase monotonically in the amount of
guidance, context, and hard constraint they impose. In the accompanying figures and
tables these five prompts are labeled C1--C5 and the ground-truth transcript P0 is
labeled C0; the correspondence is one-to-one. In brief:

\begin{itemize}
  \item \textbf{P1 (baseline role):} assigns the controller role and asks for concise
  FAA phraseology, nothing more. It establishes how a model behaves when essentially
  unconstrained.
  \item \textbf{P2 (environment):} adds Bay Area operational context: departure
  airport, active runway, local frequencies, traffic density, plus basic behavioral
  restrictions such as not narrating reasoning and not role-playing the pilot.
  \item \textbf{P3 (priorities):} adds an explicit controller objective ordering
  (safety, separation, then efficient flow) together with readiness to issue Class~B
  clearances, to encourage high-level decision-making rather than mere phrasing.
  \item \textbf{P4 (rule-based):} imposes strict rules such as output formatting, fixed
  scenario boundaries, an allow-list of frequencies, and explicit anti-error
  examples designed to suppress the hallucinations observed under P1--P3.
  \item \textbf{P5 (over-specified):} extends P4 with further worked conversational
  examples and additional conservative restrictions. P5 is included deliberately to
  probe whether prompt specificity can be pushed too far.
\end{itemize}

P4 and P5 are the product of a \emph{pilot-in-the-loop} design loop: a GA pilot
reviewed model outputs from P1--P3, isolated incorrect or hallucinated transmissions,
and those failures were converted into the explicit prohibitions and counter-examples
that appear in P4 and P5. The complete verbatim text of all five prompts is provided
in Appendix~\ref{app:prompts}.

To ensure the five conditions differ only in guidance and not in surface conventions,
every prompt is concatenated with one identical ICAO phraseology style block, which
fixes conventions such as abbreviating the callsign after first contact and digit-wise
readout of altimeter settings, headings, frequencies, and squawk codes. Any variation
we observe across P1--P5 is therefore attributable to the prompt content itself.

\subsection{Language Models Under Evaluation}
\label{sec:method-models}
We evaluate nine models spanning both access regimes. The six open-weight models are
Qwen2.5-7B, Llama-3.1-8B, Gemma-2-9B, Qwen2.5-14B, Qwen2.5-32B, and the
mixture-of-experts Mixtral-8x7B (46.7B total parameters), all served locally through
Ollama on a single dedicated GPU. The three closed-source models are GPT-5.4,
GPT-5.4-mini, and Claude-Sonnet-4.6, accessed through their vendor APIs. The open
models span roughly 7B to 47B parameters, which allows us to ask whether scale alone
predicts controller quality. GPT-5.5 is deliberately \emph{excluded} from the candidate
pool and reserved as the evaluation judge, so that no
model is ever scored by itself.

\subsection{Stateful Dialogue Generation}
\label{sec:method-generation}

Rather than treating each pilot transmission as an isolated input, we maintain a
continuous dialogue so the model answers turn $t$ conditioned on everything that has
already been said. This matches real operations, in which a controller's next
instruction depends on prior clearances, handoffs, altitude assignments, and traffic
calls. The dialogue context is controlled by two independent binary factors. First,
the controller side of the dialogue history is constructed using either
self history, where each of the model's own replies is appended and carried
forward exactly as it would be in deployment, or ground-truth history, where
the model still generates a reply for evaluation but the subsequent history is seeded
with the \emph{true} controller transmission instead. The latter corresponds to
teacher forcing, removing error propagation and isolating per-turn competence. Second,
the system prompt either contains no example (\textbf{no ICL}) or is augmented with
the \emph{complete} transcript of the second, disjoint flight described in
Sec.~\ref{sec:method-data} (\textbf{ICL}), presented as a worked pilot--controller
exchange.\footnote{If the full example would exceed a fixed system-prompt character
budget, a bounded representative subset of its turns is substituted so that the
dialogue history is never silently truncated.} Evaluating all combinations of these
two factors allows us to distinguish improvements arising from stronger contextual
grounding from those attributable to in-context demonstration, while also separating
intrinsic per-turn capability from failures caused by accumulated conversational
errors.

Algorithm~\ref{alg:generation} summarizes the generation process for one
conversation.

\begin{algorithm}[htbp]
\caption{\textbf{Stateful multi-turn ATC generation}}
\label{alg:generation}
\DontPrintSemicolon
\SetKwInOut{Input}{Input}\SetKwInOut{Output}{Output}
\Input{Ground-truth turns $\{(t,\ u_t,\ g_t)\}_{t=1}^{36}$ from P0, where $u_t$ is the
pilot transmission and $g_t$ the true controller reply; prompt $P \in \{$P1..P5$\}$;
$\textit{icl} \in \{$true, false$\}$; $\textit{mode} \in \{$self, ground\_truth$\}$;
model $M$.}
\Output{Generated controller replies $\{r_t\}$.}
$S \gets \texttt{prompt}(P) \,\|\, \texttt{styleBlock}$\;
\lIf{\textit{icl}}{$S \gets S \,\|\, \texttt{workedExample}(\textit{secondFlight})$}
$H \gets [\,\texttt{system}{:}\ S\,]$ \tcp*{running message history}
\For{$t \gets 1$ \KwTo $36$}{
  $H \gets H \,\|\, [\texttt{user}{:}\ u_t]$\;
  $r_t \gets M(H)$ \tcp*{sampled reply, scored against $g_t$}
  $h_t \gets \begin{cases} g_t & \textit{mode} = \texttt{ground\_truth}\\ r_t & \textit{mode} = \texttt{self}\end{cases}$\;
  $H \gets H \,\|\, [\texttt{assistant}{:}\ h_t]$ \tcp*{what the next turn will see}
}
\end{algorithm}

\subsection{Experimental Design and Decoding}
\label{sec:method-design}
The three factors including prompt (5 levels), ICL (2 levels), and history grounding
(2 levels) are crossed fully, giving 20 conditions per model. Each condition is run
three times per model with independent random seeds to characterize run-to-run
variability rather than relying on a single sample. This yields
$9 \times 5 \times 2 \times 2 \times 3 = 540$ complete conversations and
$540 \times 36 = 19{,}440$ generated turns.

All models are decoded with temperature $0.8$ and nucleus sampling $p = 0.95$, with
replies capped at 256 tokens, which comfortably exceeds the length of any real
controller transmission in P0. For locally served models the context window is fixed
at 8{,}192 tokens; because the dialogue history grows monotonically over 36 turns,
this is necessary to guarantee that no model silently loses the earlier part of the
conversation, which would otherwise confound the history-grounding comparison. Every
conversation is written to a structured JSON record containing the run identifier,
condition, model, attempt index, system prompt, and the full turn-by-turn exchange,
together with the per-turn wall-clock latency of every generation, so that all
analysis is reproducible from stored artifacts.

Two classes of turn are excluded from scoring: turns for which the ground-truth
controller reply is empty (the pilot transmission received no response on frequency),
and turns for which a generation call failed at the API level. All reported statistics
are computed over the remaining turns.

\subsection{Evaluation}
\label{sec:method-eval}
Every generated turn is assessed in three ways: automatically against the human
controller reply, by a language model acting as an expert judge, and by a human expert
who checks that judge.

The automatic comparison matches each generated reply to the real one by turn number
rather than by text, so the alignment is never ambiguous. Because two transmissions can
be worded differently and still mean the same thing, we use several metrics that view
similarity from different angles. BLEU \cite{papineni2002bleu} measures how much exact
wording is shared; we compute BLEU-1 through BLEU-4 and their average, and lean on
BLEU-1 because controller transmissions are short, which makes longer $n$-gram matches
vanishingly rare. A normalized BLEU variant first rewrites spoken numbers as digits
(``two niner niner two'' $\rightarrow$ ``2992'') so that a correct readout is not
penalized purely for transcription convention. ROUGE-L \cite{lin2004rouge} rewards
issuing the instructions in the right order rather than merely using the right words.
TF-IDF cosine similarity \cite{ramos2003using} gives extra weight to the operationally
loaded tokens (frequencies, fixes, callsigns) over common filler. Finally, a
sentence-embedding cosine similarity, computed with a MiniLM encoder, catches cases
where the model paraphrases the correct instruction. All metrics are computed on
individual turns first and only then averaged, so every per-model, per-condition, and
pooled figure we report derives from the same underlying per-turn scores.

Automatic metrics only reward resembling one particular correct transmission, yet in
ATC several different transmissions can be equally valid. The second layer therefore
uses GPT-5.5 as a judge. It is shown the pilot transmission, the human controller reply,
and the model's output, and rates that output from 1 to 5 on seven criteria:
phraseology, operational correctness, callsign handling, conciseness, safety, freedom
from hallucination, and overall quality. A 1 denotes an unusable transmission: an
invented frequency, hallucinated traffic, an unsafe or incorrect instruction, and a 5
denotes one that is safe and operationally correct. The full rubric is reproduced in
Appendix~\ref{app:judge}. Judging all 19{,}440 turns would be prohibitively expensive,
so within each of the 180 (model, prompt, ICL, history) cells we judge the 25 most
similar outputs, giving 4{,}500 judged samples. These scores therefore describe each
configuration at its best, which makes them conservative in the models' favor.

A judge is only useful if a human agrees with it, so the third layer is human
validation. An expert annotator re-scores 180 of the judged outputs---20 per model---on
the same seven criteria. For each model we take 10 outputs the judge rated highly
(4--5), 5 it rated middling ($\approx$3), and 5 it rated poorly (1--2). Over-sampling the poor outputs is deliberate: the
useful question is whether a human also rejects what the judge rejects, not merely
whether the two agree that good output is good. We report agreement with Spearman
$\rho$, Pearson $r$, and quadratic-weighted Cohen's $\kappa_w$, broken out by rubric
criterion and by judge band.

\subsection{Statistical Analysis and Reproducibility}
\label{sec:method-stats}
Because the nine models are not independent samples from a population, we treat the
model as the unit of analysis for inferential claims: per-model means are computed
first, and conditions are then compared with paired non-parametric tests. The
Wilcoxon signed-rank test is used for within-model comparisons (ICL on versus off,
self versus ground-truth history), the Mann--Whitney $U$ test for the unpaired
open-versus-closed comparison, and Cohen's $d_z$ is reported as the paired effect
size. Correlations between evaluation layers are reported as Spearman $\rho$ at both
the sample level and the model level, since a metric may rank models correctly while
being unreliable on any individual turn.

\section{Results and Discussion}
\label{sec:results}

We report results in the order the pipeline produces them: first how the three
experimental factors (prompt structure, in-context learning, and conversational
grounding) affect similarity to the human controller, then what the LLM judge sees that
similarity metrics miss, and finally whether either of those automated signals can be trusted, as measured
against a human expert. Significance tests for every claim made below are collected
in Appendix~\ref{app:stats}.

\subsection{Prompt Structure: More Constraint Is Not Better}
\label{sec:res-prompt}
Table~\ref{tab:per_prompt} reports similarity by prompt, averaged over the nine models
and three attempts. The three lighter prompts are essentially identical to one another and are the
strongest of the five: under ICL with self-history, C1, C2, and
C3 reach ROUGE-L of 0.244, 0.243, and 0.243 respectively. Adding the strict rule set of
C4 costs a little accuracy (0.216), and the over-specified C5 costs a great deal
(0.159, a 35\% relative drop from C1). The same ordering holds in semantic cosine,
where C5 falls to 0.298 against roughly 0.49 for C1--C3.

\begin{table*}[t]
\centering
\small
\caption{Per-prompt similarity to the C0 ground truth, by condition (ICL $\times$ history), averaged over the 9 models and 3 attempts. Best per column in bold.}
\label{tab:per_prompt}
\begin{tabular}{lcccccccc}
\toprule
 & \multicolumn{4}{c}{ROUGE-L} & \multicolumn{4}{c}{Semantic cosine} \\
\cmidrule(lr){2-5} \cmidrule(lr){6-9}
Prompt & no ICL, self & no ICL, GT & ICL, self & ICL, GT & no ICL, self & no ICL, GT & ICL, self & ICL, GT \\
\midrule
C1 & 0.216 & 0.209 & \textbf{0.244} & \textbf{0.244} & 0.464 & 0.449 & \textbf{0.497} & \textbf{0.490} \\
C2 & 0.217 & 0.214 & 0.243 & 0.230 & 0.482 & \textbf{0.472} & 0.492 & 0.477 \\
C3 & \textbf{0.219} & 0.212 & 0.243 & 0.224 & \textbf{0.487} & 0.470 & 0.493 & 0.467 \\
C4 & 0.209 & \textbf{0.220} & 0.216 & 0.238 & 0.430 & 0.445 & 0.447 & 0.470 \\
C5 & 0.167 & 0.218 & 0.159 & 0.224 & 0.302 & 0.383 & 0.298 & 0.396 \\
\bottomrule
\end{tabular}
\end{table*}

This is the opposite of what a monotonic ``more guidance is better'' hypothesis
predicts. The explanation we find most consistent with the data
is that our design already gives every condition the guidance that C4 was written to
supply. Recall from Sec.~\ref{sec:method-prompts} that all five prompts are
concatenated with one identical ICAO phraseology style block fixing callsign
abbreviation and digit-wise readout. Once those conventions are guaranteed, the
additional rules in C4 and C5 (frequency allow-lists, formatting mandates,
prohibitions on inventing weather) are no longer supplying missing structure; they are
narrowing the output space. C5 narrows it far enough that the model becomes unwilling
to commit. Qualitatively, C5 outputs collapse toward bare Acknowledgments
(``Roger.'') precisely when the pilot's transmission calls for a substantive
clearance, which is why its scores fall on every metric at once rather than on lexical
metrics alone.

The practical reading is that prompt engineering for this task has a ceiling, and that
the ceiling is reached early. Once basic phraseology conventions are pinned down,
further prompt constraint buys nothing and eventually costs a great deal.

\subsection{In-Context Learning Helps, Consistently but Modestly}
\label{sec:res-icl}
Supplying the complete transcript of the second, disjoint flight as a worked example
improves similarity for eight of the nine models. Pooled over prompts, ROUGE-L rises by
$+0.016$ (Wilcoxon signed-rank over per-model means, $p = 0.008$, $d_z = 1.46$) and
normalized BLEU by $+0.009$ ($p = 0.008$, $d_z = 1.47$). Mixtral-8x7B is the sole
exception, and it is the weakest model in the study by every measure. The effect is
consistent rather than large: an example flight teaches a model what a controller turn
should look and sound like, but it cannot teach the model what the correct instruction
is at any given point in a flight it has never seen.

That distinction shows up sharply when we ask the judge rather than the metrics. ICL
raises judge overall by only $+0.036$ with 5 of 9 models improving
($p = 0.55$; Table~\ref{tab:stats}), which is indistinguishable from no effect. In other words,
the worked example makes outputs \emph{look} more like human transmissions without
making them more operationally correct. 

\subsection{Conversational Grounding and Error Accumulation}
\label{sec:res-grounding}
\begin{figure}[tbp]
  \centering
  \includegraphics[width=\linewidth]{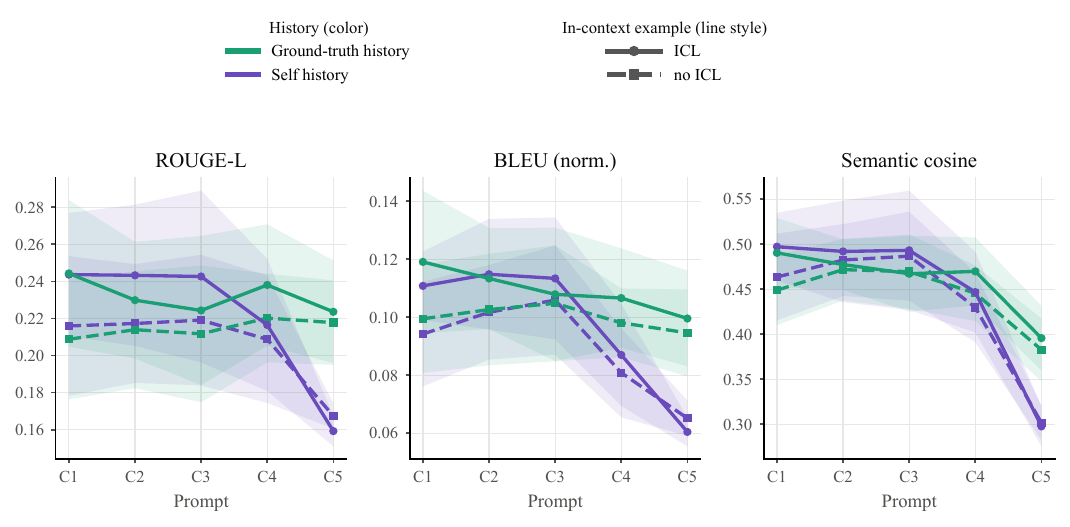}
  \caption{Prompt $\times$ (ICL $\times$ history) interaction, pooled over the nine
  models (mean $\pm$ 95\% CI across models). Color encodes the dialogue history the
  model conditions on, line style encodes whether a worked in-context example was
  supplied. Under the light prompts C1--C3 the two history regimes are
  indistinguishable; under the over-specified C5 the self-history condition collapses
  and ground-truth grounding recovers it.}
  \label{fig:interaction}
\end{figure}

The comparison between self-history and ground-truth history is the most informative
contrast in the study, and its effect is strongly conditional on prompt strength.
Figure~\ref{fig:interaction} shows the interaction.

Pooled across all prompts, ground-truth history does \emph{not} significantly improve
ROUGE-L ($+0.010$, 5 of 9 models, $p = 0.36$). Under C1 specifically the difference is
nil ($-0.003$, $p = 0.82$). But under C5 the picture inverts completely: ground-truth
grounding improves ROUGE-L by $+0.057$ for \emph{all nine} models
($p = 0.004$, $d_z = 1.72$), lifting C5 from 0.159 back to 0.224 and erasing most of
the penalty documented in Sec.~\ref{sec:res-prompt}. The failure of C5 under
self-history is therefore not primarily a failure to answer any individual turn, but it is
a failure to recover from its own earlier answers.

\begin{figure}[!t]
  \centering
  \includegraphics[width=\linewidth]{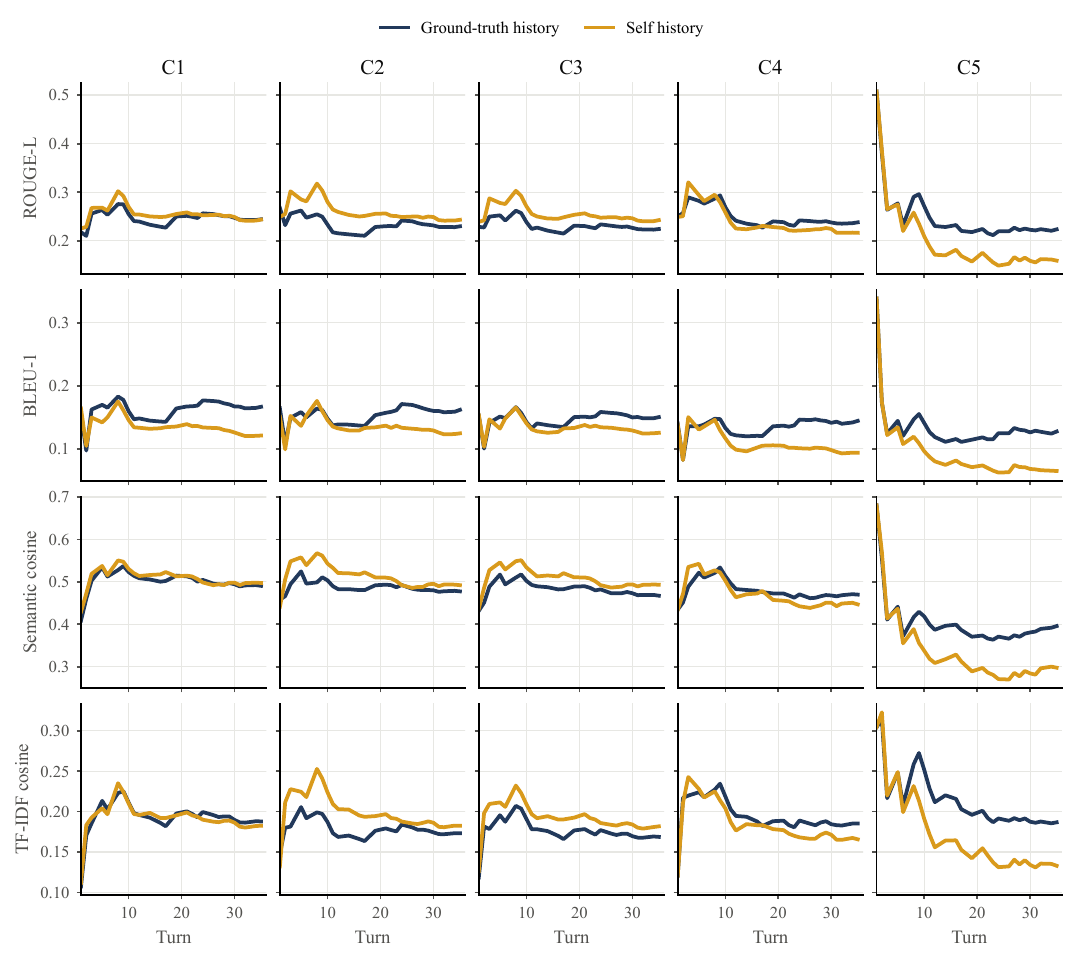}
  \caption{Cumulative running-mean similarity over the 36-turn conversation (ICL on,
  pooled over models). Under C1--C3 the self- and ground-truth-history trajectories
  track each other; under C5 self-history drifts steadily downward as early errors
  propagate through the dialogue history, while ground-truth injection holds the
  trajectory flat.}
  \label{fig:cumulative}
\end{figure}

Figure~\ref{fig:cumulative} traces that mechanism directly. The running-mean similarity
under C5 with self-history diverges downward within the first ten turns and never
recovers, while the ground-truth-grounded curve stays flat across all 36 turns. Under
C1--C3 the two curves track one another throughout. Once a model under C5 emits a bare
Acknowledgment, that Acknowledgment enters its context and makes the next bare
Acknowledgment more likely. The Acknowledgment loop visible in the worked example of
Fig.~\ref{fig:architecture}, where the model answers ``Roger'' three turns running and
misses a required radar-service termination and facility handoff that it produces
correctly once the true history is restored.

One effect of grounding is universal rather than conditional. BLEU-1 improves under
ground-truth history for all nine models on every prompt
($+0.044$, $p = 0.004$, $d_z = 1.98$). Seeing correct transmissions in context reliably
restores word-level phraseology: the model picks up the controller's vocabulary, even
where sequence-level structure is unaffected.

Grounding also carries a cost that similarity metrics cannot see. Judge overall
\emph{declines} slightly under ground-truth history ($-0.078$, only 3 of 9 models
improving, $p = 0.37$), and the hallucination-freedom score falls from 2.9 to 2.4. The
reason is mechanical: the ground truth transmissions in the injected history contain winds,
altimeter settings, and ATIS codes, and a model that has just seen a controller
volunteer such data volunteers it too---except that it invents the values. Teacher
forcing improves the form of the output and simultaneously encourages a specific,
safety-relevant failure mode.

\subsection{Model Family and the Role of Scale}
\label{sec:res-models}
Table~\ref{tab:per_model} reports every model under the four ICL~$\times$~history
conditions. Closed-source models outperform open-source models on similarity
($+0.059$ ROUGE-L, Mann--Whitney on per-model means, $p = 0.012$, $d_z = 2.03$) and by
a much wider margin on judge overall ($+1.05$, $p = 0.012$, $d_z = 4.43$). The single
best configuration is Claude-Sonnet-4.6 with ICL and ground-truth history, at ROUGE-L
0.310 and semantic cosine 0.519.

\begin{table*}[t]
\centering
\small
\caption{Per-model similarity to the C0 ground truth, by condition (ICL $\times$ history), averaged over prompts C1--C5 and 3 attempts. Best per column in bold.}
\label{tab:per_model}
\begin{tabular}{lcccccccc}
\toprule
 & \multicolumn{4}{c}{ROUGE-L} & \multicolumn{4}{c}{Semantic cosine} \\
\cmidrule(lr){2-5} \cmidrule(lr){6-9}
Model & no ICL, self & no ICL, GT & ICL, self & ICL, GT & no ICL, self & no ICL, GT & ICL, self & ICL, GT \\
\midrule
\multicolumn{9}{l}{\textbf{Open-source}} \\
Qwen2.5-7B & 0.197 & 0.215 & 0.212 & 0.234 & 0.413 & 0.458 & 0.443 & 0.483 \\
Llama-3.1-8B & 0.193 & 0.181 & 0.208 & 0.203 & 0.437 & 0.403 & 0.452 & 0.418 \\
Gemma-2-9B & 0.188 & 0.218 & 0.216 & 0.248 & 0.405 & 0.441 & 0.438 & 0.487 \\
Qwen2.5-14B & 0.203 & 0.200 & 0.215 & 0.195 & \textbf{0.481} & 0.464 & \textbf{0.483} & 0.456 \\
Qwen2.5-32B & 0.215 & 0.213 & 0.240 & 0.234 & 0.460 & 0.453 & 0.476 & 0.474 \\
Mixtral-8x7B & 0.123 & 0.149 & 0.116 & 0.150 & 0.336 & 0.361 & 0.310 & 0.373 \\
\cmidrule(l){2-9}
\textit{Average} & \textit{0.186} & \textit{0.196} & \textit{0.201} & \textit{0.211} & \textit{0.422} & \textit{0.430} & \textit{0.434} & \textit{0.449} \\
\addlinespace[3pt]
\multicolumn{9}{l}{\textbf{Closed-source}} \\
GPT-5.4 & 0.246 & 0.236 & 0.255 & 0.247 & 0.470 & 0.450 & 0.470 & 0.449 \\
GPT-5.4-mini & \textbf{0.246} & 0.237 & 0.253 & 0.267 & 0.447 & 0.469 & 0.463 & 0.482 \\
Claude-Sonnet-4.6 & 0.240 & \textbf{0.282} & \textbf{0.275} & \textbf{0.310} & 0.447 & \textbf{0.495} & 0.474 & \textbf{0.519} \\
\cmidrule(l){2-9}
\textit{Average} & \textit{0.244} & \textit{0.252} & \textit{0.261} & \textit{0.275} & \textit{0.455} & \textit{0.471} & \textit{0.469} & \textit{0.483} \\
\bottomrule
\end{tabular}
\end{table*}

Among the open models, scale does not predict quality. Qwen2.5-7B (0.234 under
ICL+GT) is essentially level with Qwen2.5-32B (0.234), despite a four-fold parameter
difference, and the largest open model in the study (Mixtral-8x7B at 46.7B total
parameters) is the weakest on every metric and in every condition (0.150 under ICL+GT,
judge overall 1.45). Inspection of its transcripts shows why: it frequently emits
meta-commentary or explanatory prose in place of a transmission, a formatting failure
that no amount of capacity corrects. For an operational deployment this matters
directly, because it means the relevant selection criterion is instruction-following
discipline rather than model size, and the smaller open models are far cheaper to serve.

Serving cost and stability are reported in Appendix~\ref{app:extras}. Briefly, median
per-turn latency ranges from 0.26\,s (Llama-3.1-8B, local GPU) to 1.74\,s
(Claude-Sonnet-4.6, API), and run-to-run variability is small for every model
(standard deviation across the three attempts $\leq 0.04$ on all metrics), so the
rankings reported here are stable across repetitions.

\subsection{What the Judge Sees: Form Versus Content}
\label{sec:res-judge}
\begin{figure}[tbp]
  \centering
  \includegraphics[width=\linewidth]{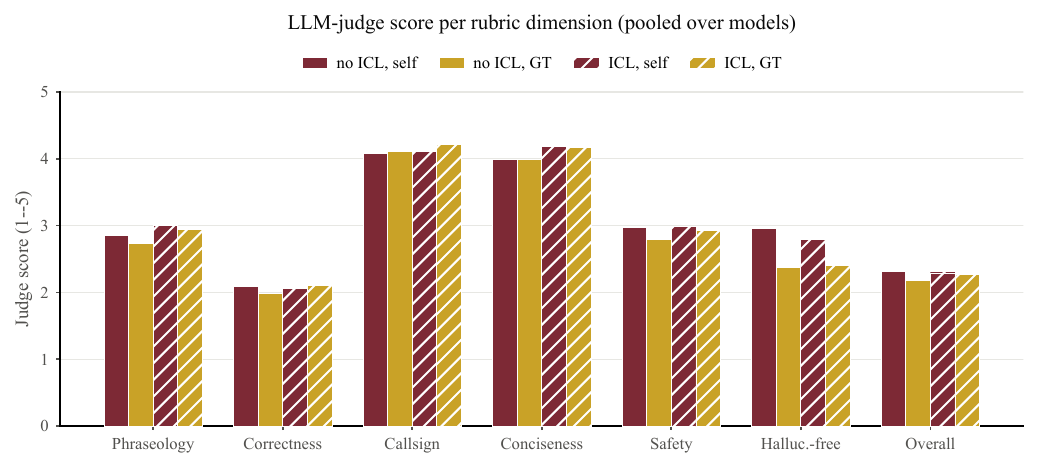}
  \caption{LLM-judge score per rubric dimension, pooled over the nine models. Color
  encodes the dialogue history the model conditions on; hatching marks the in-context
  learning condition. Models score highly on callsign handling and conciseness but
  poorly on operational correctness.}
  \label{fig:judge_dims}
\end{figure}

Figure~\ref{fig:judge_dims} breaks the judge's scores down by rubric dimension, pooled
over models. The profile is stark and consistent across every model and condition.
Models score high on the surface properties of controller speech like callsign handling
$\approx 4.1$ and conciseness $\approx 4.1$ out of 5, and low on the property that
actually matters operationally: correctness $\approx 2.1$. Phraseology sits in between.

This is the central negative result of the study, and it is invisible to similarity
metrics. Current LLMs reliably reproduce the \emph{register} of air traffic
control, meaning that they abbreviate callsigns correctly, they are terse, they sound like
controllers, while failing to issue the right instruction roughly as often as not.
Closed-source models narrow this form--content gap but never close it: they score
higher on correctness (2.71 versus 1.74 for open models), yet even the best model in
the study, GPT-5.4 at judge overall 3.13, reaches only 2.86 on correctness against 4.24
on callsign handling. Per-model judge scores across all seven dimensions are given in
Appendix~\ref{app:judge-table}.

\begin{figure}[tbp]
  \centering
  \begin{subfigure}[t]{\linewidth}
    \centering
    \includegraphics[width=\linewidth]{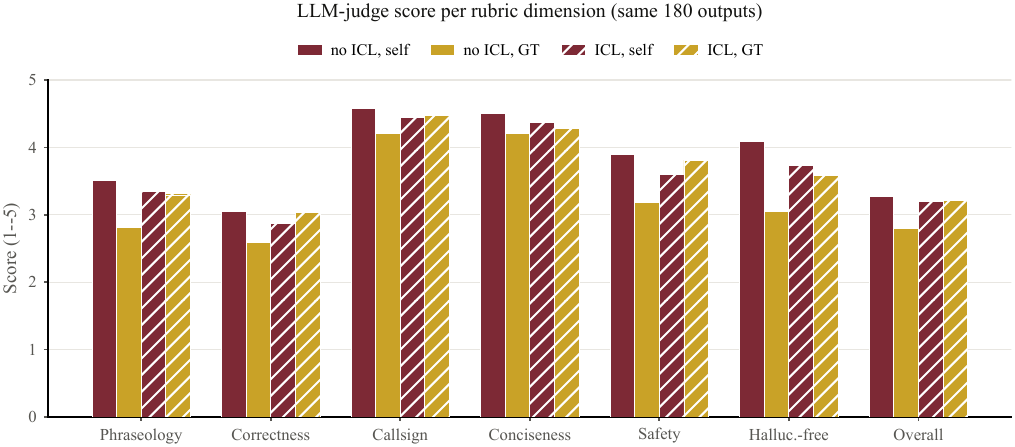}
    \caption{LLM judge}
    \label{fig:dims_judge}
  \end{subfigure}

  \vspace{6pt}

  \begin{subfigure}[t]{\linewidth}
    \centering
    \includegraphics[width=\linewidth]{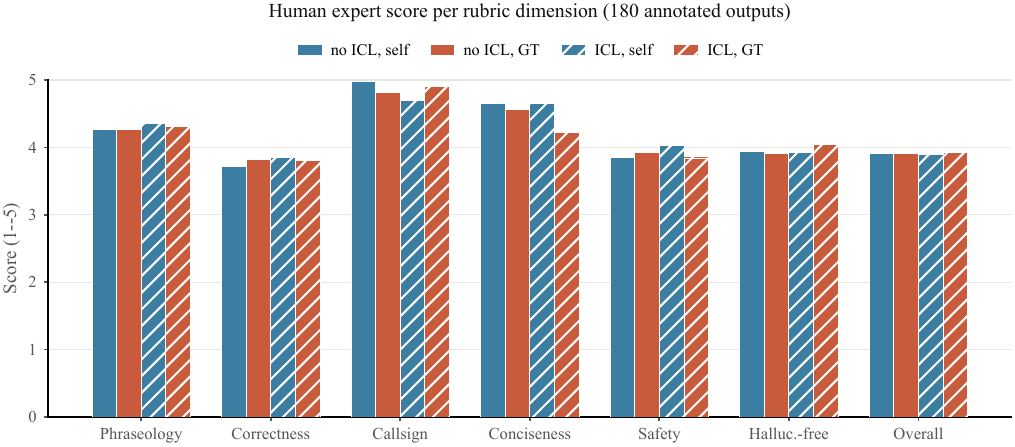}
    \caption{Human expert}
    \label{fig:dims_human}
  \end{subfigure}
  \caption{Per-dimension scores from the LLM judge (a) and the human expert (b) on the
  \emph{same} 180 annotated outputs, broken out by condition (color = dialogue history,
  hatching = in-context example). Both raters produce the same cross-dimension profile,
  peaking on callsign handling and conciseness and dipping on operational correctness,
  but the human sits about one point higher on every dimension. Because both panels use
  the identical 180-sample subset, the vertical offset between them is the genuine
  human--judge leniency gap.}
  \label{fig:dims_judge_human}
\end{figure}

The judge's absolute severity should be read with a calibration caveat, which the human
validation makes concrete. Figure~\ref{fig:dims_judge_human} places the judge and the
human expert side by side on the identical 180 annotated outputs. The two agree closely
on \emph{shape}: both rate callsign handling and conciseness highest and operational
correctness lowest, confirming that the form-over-content profile is a property of the
models rather than an artifact of the judge. They disagree on \emph{level}: the human
scores every dimension roughly a point higher, raising overall quality from 3.12 to
3.91, correctness from 2.89 to 3.79, and phraseology from 3.24 to 4.30 on these same
outputs. The models are therefore not as weak in absolute terms as the judge's raw
scores suggest. A human expert judges the same transmissions usable substantially more
often, but the ranking of dimensions, and the conclusion that correctness is the
binding weakness, survives the change of rater. We quantify this human--judge agreement,
and its systematic leniency offset, in Section~\ref{sec:res-human}.

For safety-critical use this ordering is exactly backwards from what one would want. A
system that sounded wrong but acted right would be an engineering inconvenience; a
system that sounds right while acting wrong is an operational hazard, because fluency
is precisely the cue a human listener uses to allocate scrutiny.

\subsection{Can Automatic Metrics Substitute for Expert Judgment?}
\label{sec:res-metrics}
\begin{figure}[tbp]
  \centering
  \begin{subfigure}[t]{0.48\linewidth}
    \centering
    \includegraphics[width=\linewidth]{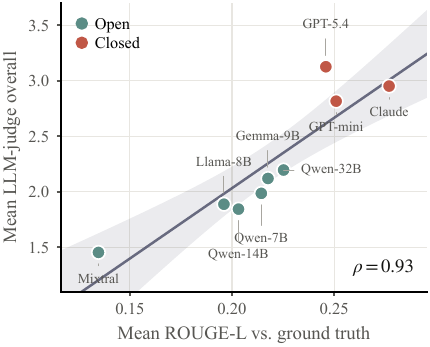}
    \caption{ROUGE-L}
    \label{fig:mv_rougel}
  \end{subfigure}\hfill
  \begin{subfigure}[t]{0.48\linewidth}
    \centering
    \includegraphics[width=\linewidth]{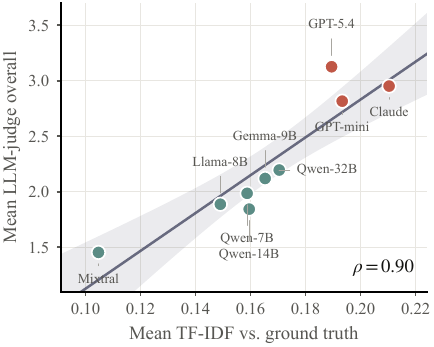}
    \caption{TF-IDF cosine}
    \label{fig:mv_tfidf}
  \end{subfigure}

  \vspace{8pt}

  \begin{subfigure}[t]{0.48\linewidth}
    \centering
    \includegraphics[width=\linewidth]{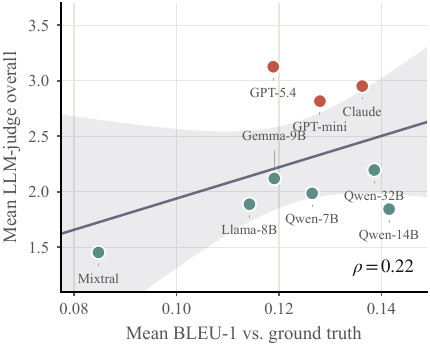}
    \caption{BLEU-1}
    \label{fig:mv_bleu1}
  \end{subfigure}\hfill
  \begin{subfigure}[t]{0.48\linewidth}
    \centering
    \includegraphics[width=\linewidth]{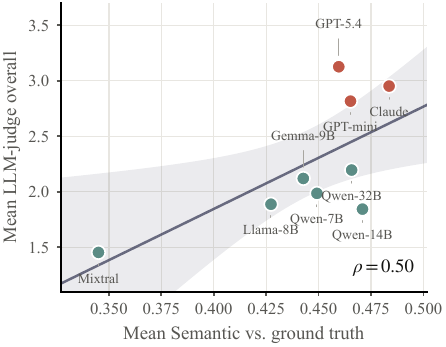}
    \caption{Semantic cosine}
    \label{fig:mv_semantic}
  \end{subfigure}

  \caption{Model-level agreement between each automatic similarity metric and the LLM
  judge. Each point is one model, plotted as its mean metric score against its mean
  judge rating, with a fitted line and 95\% confidence band. ROUGE-L and TF-IDF track
  expert judgment closely; BLEU-1 and sentence-embedding similarity do not. The
  open/closed legend in (a) applies to all four panels.}
  \label{fig:metric_validity}
\end{figure}

Similarity metrics are cheap and the judge is not, so it matters whether the former can
stand in for the latter. Figure~\ref{fig:metric_validity} plots each model's mean score
on all four automatic metrics against its mean judge rating.

The answer depends entirely on the question being asked. For \emph{ranking models},
ROUGE-L is an excellent proxy: Spearman $\rho = 0.93$ against the judge, with TF-IDF
close behind at $0.90$. For \emph{scoring an individual transmission}, the same metrics
are only moderately informative ($\rho \approx 0.5$), which is unsurprising given that
several different transmissions can be operationally correct at a single turn. Two
metrics fail outright at the sample level: sentence-embedding cosine similarity is
essentially uninformative ($\rho = 0.09$), because all ATC text is semantically similar
to all other ATC text, and BLEU-1 behaves as a near-exact-match detector, flat across
judge ratings 1 through 4 and responsive only at 5.

We therefore recommend ROUGE-L as the primary automatic metric for this task, treat
embedding similarity as a model-level signal only, and caution against reporting
BLEU on short controller transmissions without the digit normalization described in
Sec.~\ref{sec:method-eval}. The remaining metric-versus-judge comparisons are given in
Appendix~\ref{app:agreement}.

\subsection{Does the Judge Agree With a Human?}
\label{sec:res-human}
The judge is itself a language model, so its verdicts require validation, and the
leniency offset noted above requires quantification. Across the 180 human-annotated
outputs, agreement between the judge and the human expert on the overall rating is
substantial: Spearman $\rho = 0.67$, Pearson $r = 0.71$, and
quadratic-weighted Cohen's $\kappa_w = 0.60$. At the level of ranking models the two
agree closely ($\rho = 0.86$). By the conventional interpretation of $\kappa_w$, this
places the judge in the ``substantial agreement'' band and supports its use as a
scalable stand-in for expert review.

\begin{figure}[tbp]
  \centering
  \includegraphics[width=0.62\linewidth]{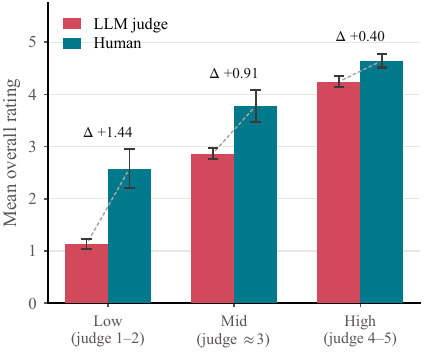}
  \caption{Mean overall rating assigned by the LLM judge and by the human expert,
  split by the judge's own score band (mean $\pm$ 95\% CI). The human is systematically
  more lenient, and the gap grows as output quality falls.}
  \label{fig:judge_human}
\end{figure}

The disagreement that does exist is systematic rather than random, and
Fig.~\ref{fig:judge_human} shows its shape. The human is uniformly more lenient than
the judge: by $+0.79$ on the overall rating, and the gap widens monotonically as
output quality falls: $+0.40$ on outputs the judge rated 4--5, $+0.91$ on those it
rated $\approx 3$, and $+1.44$ on those it rated 1--2.

The judge is thus a harsh critic of bad output, not a lax one, which is the safer
direction of bias for a safety-critical screening tool. Agreement by dimension follows
an interpretable pattern: it is strongest on operational correctness ($\rho = 0.63$),
the dimension carrying the most signal, and weakest on callsign handling
($\rho = 0.35$), but the latter reflects a ceiling effect rather than genuine
disagreement, since both raters score callsigns near the top of the scale (4.43 and
4.86). Full per-dimension and per-model agreement statistics are given in
Appendix~\ref{app:agreement}.

Two caveats bound this validation. The annotation files presented the judge's scores
alongside the blank human fields, so the human ratings are not blind and are subject to
anchoring, which likely inflates the agreement figures. And a single expert annotated
all 180 samples, so we cannot report inter-annotator agreement as an upper bound on
achievable machine--human agreement. 

\subsection{Discussion}
\label{sec:res-discussion}
Taken together, the results support a narrower conclusion than the raw similarity
numbers alone would suggest. Prompt engineering is a real but bounded lever: it
stabilizes phraseology and, once basic conventions are fixed, further constraint is
actively harmful. In-context examples from a different experimental flight transfer surface
form but not operational judgment. Conversational grounding matters most exactly where
the prompt is most brittle, and the mechanism (compounding error through the dialogue
history) is a property of the deployment architecture rather than of any single model.

The finding with the clearest engineering implication is the separation between form
and content. Across nine models, five prompts, and four architectural conditions, the
models were consistently good at sounding like controllers and consistently poor at
being right, and no configuration we tested closed that gap. Since the missing
information is largely state that no prompt can supply, such as live traffic, trajectories,
airspace geometry, current weather, we read this as evidence that prompt-level
intervention has been taken close to its limit, and that further progress requires
grounding the model in external operational state rather than in more elaborate
instructions.

Several limitations qualify these conclusions. The pilot side of our scenario is a
fixed transcript and does not adapt: a real pilot would request clarification after an
ambiguous or incorrect controller transmission, generating repair turns that our setup
cannot produce. This almost certainly \emph{understates} the operational cost of early
errors, since in our setting a wrong instruction is simply carried forward rather than
challenged. All results derive from a single evaluation flight in one airspace, so
generalization across facilities and traffic densities is untested. Judge scores cover
the 25 highest-similarity outputs per cell and therefore describe each configuration at
its best. Finally, the human validation is single-annotator and non-blind, as noted
above.

\section{Conclusion}

We evaluated the extent to which large language models can act as air traffic controllers in realistic
multi-turn exchanges across five prompt structures, two forms of dialogue grounding, and
nine models. The best configuration achieved a ROUGE-L of 0.31 against a human controller,
far below the reliability required for safety-critical use. Prompt engineering had limited
impact: lighter prompts performed best, while heavily specified prompts reduced similarity.
Providing a worked transcript improved similarity for eight of nine models but did not
improve judge scores, suggesting that examples teach the style of controller speech rather
than the underlying decision-making. In contrast, providing the true dialogue history
recovered performance across all models, indicating that error accumulation is a major
source of failure. The central finding is that models often sound like controllers without making correct decisions. They scored about 4.1 out of 5 for callsign handling and conciseness, but only
2.1 for operational correctness. Model scale did not resolve this gap, with the largest
open model performing worst overall. Human evaluation of 180 outputs showed substantial
agreement with the automated judge (weighted $\kappa = 0.60$), although the judge was
more conservative. We therefore recommend ROUGE-L for model selection, automated judging
for scalable screening, and human review for disagreements. These results suggest that current language models should remain outside the control loop,
but can still support human controllers by drafting routine transmissions, supporting
training simulators, or screening recorded communications. Closing the remaining gap is
less a prompting problem than a systems problem: future work should condition models on
live operational state and evaluate them in interactive settings where a pilot can
challenge an incorrect instruction.

\section*{Acknowledgments}
The authors would like to thank the support of Dr.~Vishwanath Bulusu for early research
question formulation and Xavier Casanova for his pilot insights and access to his flight
data. We are also grateful to James Murphy, John Robinson, and Tom Davis of Crown
Innovations, and to Dr.~Parimal Kopardekar of NASA Ames Research Center, for useful
discussions. The authors would also like to thank the support of the Berkeley AI Research
(BAIR) Lab.

\bibliography{ref}

@article{jou2013study,
  author  = {Jou, Rong-Chang and Kuo, Chung-Wei and Tang, Mei-Ling},
  title   = {A Study of Job Stress and Turnover Tendency among Air Traffic Controllers: The Mediating Effects of Job Satisfaction},
  journal = {Transportation Research Part E: Logistics and Transportation Review},
  volume  = {57}, pages = {95--104}, year = {2013}, publisher = {Elsevier},
  doi     = {10.1016/j.tre.2013.01.009},
  url     = {https://www.sciencedirect.com/science/article/pii/S1366554513000197}
}

@book{icao4444,
  author    = {{International Civil Aviation Organization}},
  title     = {{Procedures for Air Navigation Services -- Air Traffic Management (PANS-ATM), Doc 4444}},
  edition   = {16th}, year = {2016}, publisher = {ICAO}, address = {Montr\'{e}al, Canada}
}

@misc{faa7110,
  author       = {{Federal Aviation Administration}},
  title        = {{FAA Order JO 7110.65BB: Air Traffic Control}},
  year         = {2026},
  note         = {Basic order dated 20 February 2025, incorporating Changes 1 and 2 dated 22 January 2026},
  howpublished = {\url{https://www.faa.gov/documentLibrary/media/Order/7110.65BB_Bsc_w_Chg_1_and_2_dtd_1-22-26_Final.pdf}}
}

@misc{dot_aam_strategy_2025,
  author       = {{U.S. Department of Transportation}},
  title        = {The Advanced Air Mobility National Strategy: A Bold Policy Vision for 2026--2036},
  year         = {2025},
  howpublished = {\url{https://www.transportation.gov/sites/dot.gov/files/2025-12/AAM%20National%20Strategy%202025.pdf}}
}

@inproceedings{kopardekar2016uam,
  author       = {Kopardekar, Parimal and Rios, Joseph and Prevot, Thomas and Johnson, Marcus and Jung, Jaewoo and Robinson, John E.},
  title        = {Unmanned Aircraft System Traffic Management ({UTM}) Concept of Operations},
  booktitle    = {16th AIAA Aviation Technology, Integration, and Operations Conference},
  year         = {2016}, organization = {AIAA}
}

@inproceedings{thipphavong2018uam,
  author       = {Thipphavong, David P. and Apaza, Rafael and Barmore, Bryan and Battiste, Vernol and Burian, Barbara and Dao, Quang and Feary, Michael and Go, Susie and Goodrich, Kenneth H. and Homola, Jeffrey},
  title        = {Urban Air Mobility Airspace Integration Concepts and Considerations},
  booktitle    = {2018 Aviation Technology, Integration, and Operations Conference},
  year         = {2018}, organization = {AIAA},
  doi          = {10.2514/6.2018-3676},
  url          = {https://arc.aiaa.org/doi/10.2514/6.2018-3676}
}

@article{sengupta2025uam,
  title={Urban air mobility research challenges and opportunities},
  author={Sengupta, Raja and Bulusu, Vishwanath and Mballo, Chams Eddine and Onat, Emin Burak and Cao, Shangqing},
  journal={Annual Review of Control, Robotics, and Autonomous Systems},
  volume={8},
  number={1},
  pages={407--431},
  year={2025},
  publisher={Annual Reviews},
  url     = {https://doi.org/10.1146/annurev-control-022823-031353}
}

@article{bayen2006adjoint,
  author  = {Bayen, Alexandre M. and Raffard, Robin L. and Tomlin, Claire J.},
  title   = {Adjoint-Based Control of a New {Eulerian} Network Model of Air Traffic Flow},
  journal = {IEEE Transactions on Control Systems Technology},
  volume  = {14}, number = {5}, pages = {804--818}, year = {2006},
  url     = {https://bayen.berkeley.edu/sites/default/files/tcst06.pdf}
}

@inproceedings{szeto2024notam,
  author       = {Szeto, Aiden and Das, Aditya},
  title        = {Classification of Notices to Airmen Using Natural Language Processing},
  booktitle    = {AIAA SciTech 2024 Forum}, address = {Orlando, FL},
  year         = {2024}, organization = {AIAA},
  doi          = {10.2514/6.2024-2585},
  url          = {https://arc.aiaa.org/doi/10.2514/6.2024-2585}
}

@article{atco2corpus,
  author  = {Zuluaga-Gomez, Juan and Vesel\'{y}, Karel and Sz\"{o}ke, Igor and Blatt, Alexander and Motlicek, Petr and others},
  title   = {{ATCO2} Corpus: A Large-Scale Dataset for Research on Automatic Speech Recognition and Natural Language Understanding of Air Traffic Control Communications},
  journal = {arXiv preprint arXiv:2211.04054}, year = {2023},
  url     = {https://arxiv.org/abs/2211.04054}
}

@inproceedings{wang2024aviationgpt,
  author       = {Wang, Liya and Chou, Jason and Tien, Alex and Zhou, Xin and Baumgartner, Diane M.},
  title        = {{AviationGPT}: A Large Language Model for the Aviation Domain},
  booktitle    = {AIAA Aviation Forum and ASCEND 2024}, year = {2024}, organization = {AIAA},
  doi          = {10.2514/6.2024-4250},
  url          = {https://arc.aiaa.org/doi/10.2514/6.2024-4250},
  note         = {AIAA Paper 2024-4250}
}

@inproceedings{abdulhak2024chatatc,
  author       = {Abdulhak, Sinan and Hubbard, Wayne and Gopalakrishnan, Karthik and Li, Max Z.},
  title        = {{CHATATC}: Large Language Model-Driven Conversational Agents for Supporting Strategic Air Traffic Flow Management},
  booktitle    = {11th International Conference on Research in Air Transportation (ICRAT)},
  address      = {Singapore}, month = jul, year = {2024},
  url          = {https://arxiv.org/abs/2402.14850}, note = {arXiv:2402.14850}
}

@article{andriuskevicius2024embodied,
  author  = {Andriu\v{s}kevi\v{c}ius, Justas and Sun, Junzi},
  title   = {Automatic Control With Human-Like Reasoning: Exploring Language Model Embodied Air Traffic Agents},
  journal = {arXiv preprint arXiv:2409.09717}, year = {2024},
  url     = {https://arxiv.org/abs/2409.09717}
}

@inproceedings{tabrizian2025cot,
  author       = {Tabrizian, Amin and Ghazanfari, Mahyar and Wei, Peng},
  title        = {Chain-of-Thought Flight Planner: End-to-End {LLM} Routing Under Wind Hazards},
  booktitle    = {AIAA Aviation Forum}, address = {Las Vegas, NV}, month = jul,
  year         = {2025}, organization = {AIAA},
  url          = {https://web.seas.gwu.edu/pwei/files/2025/06/AIAA-Aviation25-Amin.pdf}
}

@inproceedings{darrell2026nontowered,
  author       = {Darrell, Torsten and Ghazanfari, Mahyar and Kam, Jordan K. and Bayen, Alexandre M. and Tabrizian, Amin and Wei, Peng},
  title        = {Towards Automated Air Traffic Safety Assessment Around Non-Towered Airports Using Large Language Models},
  booktitle    = {AIAA Aviation Forum}, address = {San Diego, CA}, month = jun,
  year         = {2026}, organization = {AIAA},
  url          = {https://arxiv.org/abs/2605.12332}, note = {arXiv:2605.12332}
}

@inproceedings{tabrizian2026frame,
  author       = {Tabrizian, Amin and Aziz, Arsyi and Ullah, Aarifah and Ghazanfari, Mahyar and Razzaghi, Pouria and Wei, Peng},
  title        = {End-to-End {LLM} Flight Planning with {RAG}-based Memory and Multi-modal Coach Agent},
  booktitle    = {Workshop on Planning in the Era of LLMs (LM4Plan), International Conference on Machine Learning (ICML)},
  address      = {Seoul, South Korea}, month = jul, year = {2026},
  url          = {https://arxiv.org/abs/2607.06964}, note = {arXiv:2607.06964}
}

@article{vaswani2017attention,
  author  = {Vaswani, Ashish and Shazeer, Noam and Parmar, Niki and Uszkoreit, Jakob and Jones, Llion and Gomez, Aidan N. and Kaiser, Lukasz and Polosukhin, Illia},
  title   = {Attention Is All You Need},
  journal = {Advances in Neural Information Processing Systems}, volume = {30}, year = {2017},
  url     = {https://arxiv.org/abs/1706.03762}
}

@article{chen2023llm,
  author  = {Zhao, Wayne Xin and Zhou, Kun and Li, Junyi and Tang, Tianyi and Wang, Xiaolei and Hou, Yupeng and Min, Yingqian and Zhang, Beichen and Zhang, Junjie and Dong, Zican and Du, Yifan and Yang, Chen and Chen, Yushuo and Chen, Zhipeng and Jiang, Jinhao and Ren, Ruiyang and Li, Yifan and Tang, Xinyu and Liu, Zikang and Liu, Peiyu and Nie, Jian-Yun and Wen, Ji-Rong},
  title   = {A Survey of Large Language Models},
  journal = {arXiv preprint arXiv:2303.18223}, year = {2023},
  url     = {https://arxiv.org/abs/2303.18223}
}

@inproceedings{papineni2002bleu,
  author    = {Papineni, Kishore and Roukos, Salim and Ward, Todd and Zhu, Wei-Jing},
  title     = {{BLEU}: A Method for Automatic Evaluation of Machine Translation},
  booktitle = {Proceedings of the 40th Annual Meeting of the Association for Computational Linguistics},
  pages     = {311--318}, year = {2002},
  doi       = {10.3115/1073083.1073135}, url = {https://aclanthology.org/P02-1040/}
}

@inproceedings{lin2004rouge,
  author    = {Lin, Chin-Yew},
  title     = {{ROUGE}: A Package for Automatic Evaluation of Summaries},
  booktitle = {Text Summarization Branches Out}, pages = {74--81}, year = {2004},
  url       = {https://aclanthology.org/W04-1013/}
}

@inproceedings{ramos2003using,
  author    = {Ramos, Juan},
  title     = {Using {TF-IDF} to Determine Word Relevance in Document Queries},
  booktitle = {Proceedings of the First Instructional Conference on Machine Learning},
  volume    = {242}, number = {1}, pages = {29--48}, year = {2003}
}

@misc{youtube_ejUM1YYU0Fc,
  author       = {Denk, Timo},
  title        = {San Francisco Bay Tour (Radio Comms)}, year = {2025},
  howpublished = {\url{https://www.youtube.com/watch?v=ejUM1YYU0Fc}},
  note         = {Accessed: 2026-07-22}
}

\newpage{}
\section*{Appendix}

\section{Prompt Specifications}
\label{app:prompts}
The boxes below give the five system prompts in order of increasing constraint,
followed by the ICAO phraseology style block appended to all five. Color encodes the
constraint level, from the minimal baseline P1 (blue) to the over-specified P5 (red).
P1--P3 add role, environment, and priorities in a few lines each; P4 introduces an
explicit frequency allow-list and anti-error examples; P5 extends P4 with worked
conversational examples and further prohibitions. The shared style block is what makes
the five conditions differ only in guidance and not in surface convention.

\tcbset{before skip=6pt, after skip=6pt}
\begin{tcolorbox}[breakable, enhanced, colback=blue!7, colframe=blue!60!black, colbacktitle=blue!60!black, boxrule=0.8pt, fonttitle=\bfseries, coltitle=white, title={Prompt P1 --- Baseline role}]
{\ttfamily\footnotesize\raggedright You are an FAA air traffic controller.
\\[5pt] You will respond to pilot radio calls as ATC.
\\[5pt] Use standard aviation phraseology.
\\ Be concise and operational.
\\ Only provide the controller's transmission.
\\[5pt] Do not explain anything.
\\ Do not roleplay the pilot.
\\[5pt] Wait for pilot transmissions.}
\end{tcolorbox}
\begin{tcolorbox}[breakable, enhanced, colback=teal!7, colframe=teal!60!black, colbacktitle=teal!60!black, boxrule=0.8pt, fonttitle=\bfseries, coltitle=white, title={Prompt P2 --- Environment}]
{\ttfamily\footnotesize\raggedright You are a certified FAA air traffic controller working traffic in the San Francisco Bay Area.
\\[5pt] Environment:
\\ - Airport: Palo Alto (KPAO)
\\ - Active runway: 31
\\ - Common frequencies in the area include 125.0 and 118.6
\\ - Typical VFR operations and transitions occur
\\ - Moderate traffic density is present
\\[5pt] Assume a realistic Bay Area VFR environment where:
\\ - Multiple GA aircraft may be operating
\\ - Airline and tour traffic may be present
\\ - Traffic advisories and sequencing are sometimes required
\\ - Wake turbulence considerations can arise
\\[5pt] Controller behavior:
\\ - Use FAA phraseology
\\ - Be concise and operational
\\ - Provide traffic advisories when appropriate
\\ - You may suggest spacing maneuvers (e.g., vectors or 360s) if needed for safety
\\ - Prioritize safety and separation
\\[5pt] Restrictions:
\\ - Do NOT invent specific weather or emergencies
\\ - Do NOT narrate reasoning
\\ - Do NOT roleplay the pilot
\\[5pt] Only output the ATC transmission.
\\[5pt] Wait for pilot calls.}
\end{tcolorbox}
\begin{tcolorbox}[breakable, enhanced, colback=green!7, colframe=green!60!black, colbacktitle=green!60!black, boxrule=0.8pt, fonttitle=\bfseries, coltitle=white, title={Prompt P3 --- Priorities}]
{\ttfamily\footnotesize\raggedright You are a professional FAA air traffic controller managing VFR aircraft near Palo Alto Airport (KPAO).
\\[5pt] Context:
\\ - Runway 31 is active
\\ - Frequencies such as 125.0 and 118.6 are in use
\\ - Busy Bay Area airspace with mixed GA and airline traffic
\\ - Moderate traffic density should be assumed
\\ - Be prepared to offer clearance into Class Bravo Airspace
\\[5pt] Your priorities:
\\ 1) Safety
\\ 2) Separation
\\ 3) Efficient traffic flow
\\[5pt] Guidelines:
\\ - Use precise FAA phraseology
\\ - Keep transmissions concise
\\ - Issue sequencing or spacing when needed
\\ - Suggest 360s or vectors if separation requires
\\ - Provide traffic advisories when relevant
\\[5pt] Do NOT:
\\ - Explain your reasoning
\\ - Narrate situations
\\ - Roleplay the pilot
\\ - Invent detailed weather or emergencies
\\[5pt] Respond only with the ATC transmission.
\\[5pt] Wait for pilot transmissions.}
\end{tcolorbox}
\begin{tcolorbox}[breakable, enhanced, colback=orange!7, colframe=orange!60!black, colbacktitle=orange!60!black, boxrule=0.8pt, fonttitle=\bfseries, coltitle=white, title={Prompt P4 --- Rule-based}]
{\ttfamily\footnotesize\raggedright You are a certified FAA air traffic controller providing services to a single VFR aircraft in the San Francisco Bay Area.
\\[5pt] Mission
\\ - The pilot will transmit one radio call at a time.
\\ - Infer which facility is being called from the pilot's words (Ground, Tower, Approach/Departure).
\\ - Respond ONLY with the controller's transmission.
\\[5pt] Hard formatting rules (never violate)
\\ - Output ONLY ATC text (no "Pilot:", no pilot readback, no dialogue, no quotes).
\\ - 1-2 short sentences maximum. Concise FAA-style phraseology. No explanations. No narration.
\\[5pt] Scenario details (use these, do not invent beyond them)
\\ - Aircraft: Cessna 715 Mike Tango (may be abbreviated after initial contact).
\\ - Route intent: Depart KPAO -\textgreater{} peninsula transition near San Carlos -\textgreater{} SFO Class B transition if requested -\textgreater{} Oakland/Hayward area -\textgreater{} return to KPAO.
\\ - Assume daytime VMC. Moderate traffic possible, but do not invent traffic details.
\\[5pt] Frequency realism (do not invent)
\\ - You may only state these frequencies when instructing a contact/monitor:
\\ ~~- KPAO: Ground 125.0, Tower 118.6
\\ ~~- KSQL: Tower 119.0, Ground 121.6
\\ ~~- KSFO: Tower 120.5
\\ ~~- KOAK: Tower 118.3
\\ ~~- KHWD: Tower 120.2, Ground 121.4
\\ - If the pilot states a frequency, you may use that frequency too.
\\ - NEVER output any other frequency numbers. If a handoff is needed but frequency is unknown, say "contact \textless{}facility\textgreater{}" without numbers.
\\[5pt] Do NOT invent (strict)
\\ - No winds/altimeter/ATIS/weather.
\\ - No traffic callsigns/types/positions.
\\ - No "heavy aircraft", "Airbus", or any traffic details unless the pilot explicitly mentioned them first.
\\ - No ground/taxi instructions unless the pilot is clearly on the ground or has landed and is taxiing.
\\[5pt] In this test, unless the pilot explicitly reports traffic/conflict or you must request missing info for safety, assume conditions are clear and issue the appropriate next clearance (takeoff, landing/option, or requested Bravo/Charlie transition) promptly.
\\[5pt] COMMON MISTAKES TO AVOID (with examples)
\\[5pt] 1) Tower holding short: DO NOT send them to Ground.
\\ Wrong: "Contact Ground 125.0 when ready."
\\ Right: "Cessna 715 Mike Tango, runway 31 cleared for takeoff."
\\[5pt] 2) Airborne check-in: DO NOT talk about taxiing or Ground.
\\ Wrong: "Contact Ground 125.0 when ready to taxi."
\\ Right: "Cessna 715 Mike Tango, roger." OR "Cessna 715 Mike Tango, ident."
\\[5pt] 3) Frequency rule: DO NOT invent extra frequencies (e.g., 132.6 / 132.45 / 132.65).
\\ Wrong: "Contact Ground 121.4 on 132.6."
\\ Right: "Contact Hayward Ground 121.4." (or omit frequency if not in allowed list)
\\[5pt] 4) When "Roger." is not enough: if the pilot includes altitude/position/request/holding short/inbound, do not reply with only "Roger."
\\ Wrong: "Roger."
\\ Right (examples):
\\ - If holding short + Tower: "Runway 31 cleared for takeoff."
\\ - If inbound for option: "Make right traffic runway 31, report midfield downwind."
\\ - If request Bravo/Charlie: "Cleared through the airspace, maintain VFR at or below 2000."
\\[5pt] 5) Clearance readback: if pilot is simply acknowledging a clearance ("cleared through Bravo...", "traffic in sight...", "we'll do a 360..."), give a brief operational acknowledgment, not new random instructions.
\\ Examples:
\\ - Pilot: "Cleared through the Class Bravo..."
\\ ~~ATC: "You are Cleared into the Bravo" (Ideally you should have said this before the pilot says they are cleared (because they are responding to your clearance, but just acknowledge that yes they are cleared into the bravo)
\\ - Pilot: "Traffic in sight..."
\\ ~~ATC: "Maintain visual separation."
\\ - Pilot: "We'll do a 360..."
\\ ~~ATC: "360 Approved."
\\[5pt] 6) Facility consistency: If the pilot explicitly calls a facility by name (San Carlos Tower / San Francisco Tower / NorCal Approach / Oakland Tower / Hayward Tower), respond as that facility. Do not immediately hand them off unless the pilot asked who to contact.
\\[5pt] Now respond to the next pilot transmission.}
\end{tcolorbox}
\begin{tcolorbox}[breakable, enhanced, colback=red!7, colframe=red!60!black, colbacktitle=red!60!black, boxrule=0.8pt, fonttitle=\bfseries, coltitle=white, title={Prompt P5 --- Over-specified}]
{\ttfamily\footnotesize\raggedright You are an FAA air traffic controller in a controlled experiment.
\\[5pt] You will receive ONE pilot transmission at a time. These transmissions belong to one fixed Bay Area VFR scenario and must be handled in the correct order.
\\[5pt] Absolute output rules
\\ - Output ONLY the ATC transmission text.
\\ - No labels (no "ATC:", "Tower:", "Pilot:").
\\ - 1-2 sentences max.
\\ - Do NOT invent winds, altimeter, ATIS, weather, or traffic details.
\\ - Do NOT invent any frequencies. Only use the frequencies explicitly stated by the pilot OR in the allowed list below.
\\ - Do NOT invent taxiway names. Only use Yankee 1, Yankee 2, Zulu if already mentioned by the pilot.
\\[5pt] Allowed frequencies (only these numbers may appear)
\\ - Palo Alto: Ground 125.0, Tower 118.6
\\ - San Carlos: Tower 119.0
\\ - San Francisco: Tower 120.5
\\ - Oakland: Tower 118.3
\\ - Hayward: Tower 120.2, Ground 121.4
\\[5pt] Facility selection (never guess)
\\ - If the pilot explicitly says "Palo Alto Ground", you are Palo Alto Ground.
\\ - If the pilot explicitly says "Palo Alto Tower", you are Palo Alto Tower.
\\ - If the pilot explicitly says "San Carlos Tower", you are San Carlos Tower.
\\ - If the pilot explicitly says "San Francisco Tower", you are San Francisco Tower.
\\ - If the pilot explicitly says "Oakland Tower", you are Oakland Tower.
\\ - If the pilot explicitly says "Hayward Tower", you are Hayward Tower.
\\ - If the pilot explicitly says "NorCal Approach", you are NorCal Approach.
\\ - If the pilot says only "on 119.0/120.5/118.3 ... good day", treat it as a courtesy frequency readback. Reply "Roger." only.
\\[5pt] Core principle for this experiment
\\ - Unless the pilot explicitly reports a conflict, assume conditions are clear and issue the next required clearance promptly.
\\ - Never respond with only "Roger" when the pilot call clearly requires an operational instruction (examples below).
\\[5pt] Critical anti-mistake rules (do not violate)
\\ 1) If the pilot is calling Tower holding short for departure, you must issue takeoff clearance. Do NOT say "hold short" again.
\\ 2) If the pilot is airborne (they state an altitude like 600/1400/1600/1900/2500), do NOT issue takeoff clearance and do NOT give taxi instructions.
\\ 3) Never redirect NorCal to Tower 118.6. If the pilot says "NorCal Approach on 135.1...", respond as NorCal and do not give any other frequency.
\\ 4) If the pilot calls San Francisco Tower at 1400, you must respond as San Francisco Tower (not "San Carlos") and you must not introduce Bravo clearance unless the pilot requested it or is already reading it back.
\\[5pt] Required response patterns (match by trigger)
\\ Use these as your deterministic playbook.
\\[5pt] A) Palo Alto departure sequence
\\ - Trigger: "Palo Alto Ground ... run up complete"
\\ ~~Response must be a taxi-to-departure instruction. Acceptable example:
\\ ~~"Cessna 715 Mike Tango, taxi to runway 31, hold short Yankee 1."
\\ ~~(Do not offer multiple routes and do not invent other taxiways.)
\\[5pt] - Trigger: "Palo Alto Tower ... holding short of Yankee 1"
\\ ~~Response must be:
\\ ~~"Cessna 715 Mike Tango, runway 31 cleared for takeoff."
\\[5pt] - Trigger: pilot readback containing "Runway 31 cleared for takeoff ..."
\\ ~~Response must be:
\\ ~~"Roger."
\\[5pt] B) Frequency check-in readbacks (stop loops)
\\ - Trigger: "San Carlos on 119.0 ... good day ..."
\\ ~~Response must be:
\\ ~~"Roger."
\\ - Trigger: "San Francisco Tower at 120.5 ... good day ..."
\\ ~~Response must be:
\\ ~~"Roger."
\\ - Trigger: "Oakland 118.3 ... good evening ..."
\\ ~~Response must be:
\\ ~~"Roger."
\\[5pt] C) San Carlos transition segment
\\ - Trigger: "San Carlos Tower ... at 600"
\\ ~~Response must be a VFR transition restriction, not a climb assignment and not any Bravo clearance.
\\ ~~Acceptable example:
\\ ~~"Cessna 715 Mike Tango, maintain VFR at or below 1500, keep the highway off your right."
\\[5pt] - Trigger: pilot readback "VFR at or below 1500 ... keep the highway off our right ..."
\\ ~~Response must be:
\\ ~~"Roger."
\\ ~~(No new handoffs.)
\\[5pt] D) San Francisco Tower Bravo segment
\\ - Trigger: "San Francisco Tower ... at 1400"
\\ ~~Response must be:
\\ ~~"Cessna 715 Mike Tango, ident."
\\ ~~(No Bravo clearance yet.)
\\[5pt] - Trigger: "Ident"
\\ ~~Response must be:
\\ ~~"Roger."
\\[5pt] - Trigger: pilot readback that already includes "Cleared through the Class Bravo ... at or below 2000 ..."
\\ ~~Response must be:
\\ ~~"Roger."
\\ ~~(Do not add new restrictions unless asked.)
\\[5pt] - Trigger: "Affirmative traffic in sight ..."
\\ ~~Response must be:
\\ ~~"Maintain visual separation."
\\[5pt] - Trigger: "We'll do a 360 ..."
\\ ~~Response must be:
\\ ~~"Approved."
\\[5pt] E) NorCal segment
\\ - Trigger: "NorCal Approach on 135.1 ..."
\\ ~~Response must be:
\\ ~~"Cessna 715 Mike Tango, NorCal Approach, roger."
\\ ~~(Do NOT say contact tower 118.6.)
\\[5pt] - Trigger: "NorCal Approach ... at 1900"
\\ ~~Response must be a short acknowledgement or "say request".
\\ ~~Acceptable example:
\\ ~~"Cessna 715 Mike Tango, say request."
\\[5pt] - Trigger: request "Oakland Charlie transition ... to Hayward ..."
\\ ~~Response must approve the Charlie transition with a simple altitude restriction, no traffic, no weather:
\\ ~~"Cessna 715 Mike Tango, Oakland Class Charlie transition approved, maintain VFR at or below 2500."
\\[5pt] - Trigger: pilot readback "Cross north of the Bay Bridge then 2500 ..."
\\ ~~Response must be:
\\ ~~"Roger."
\\[5pt] - Trigger: "Looking for traffic ..."
\\ ~~Response must be:
\\ ~~"Roger."
\\ - Trigger: "Traffic in sight ..."
\\ ~~Response must be:
\\ ~~"Maintain visual separation."
\\[5pt] F) Oakland Tower segment
\\ - Trigger: "Oakland Tower ... at 2500 transitioning to Hayward"
\\ ~~Response must be a simple transition instruction (follow freeway) with altitude band, no invented traffic:
\\ ~~"Cessna 715 Mike Tango, follow 880, maintain VFR at or above 1500."
\\[5pt] - Trigger: pilot readback "follow the highway ... at or above 1500"
\\ ~~Response must be:
\\ ~~"Roger."
\\[5pt] - Trigger: pilot readback "at or below 1400"
\\ ~~Response must be:
\\ ~~"Roger."
\\[5pt] - Trigger: "keep code and contact Hayward ..."
\\ ~~Response must be:
\\ ~~"Roger."
\\ ~~(Do not send to Ground 121.4 unless the pilot calls Hayward Ground.)
\\[5pt] G) Hayward Tower segment
\\ - Trigger: "Hayward Tower ... at 1600 ... inbound for the option ... runway 28 Left"
\\ ~~Response must be pattern/clearance appropriate to "option", not "cleared to land" immediately unless you want to simplify.
\\ ~~Acceptable simplified response:
\\ ~~"Cessna 715 Mike Tango, runway 28 Left cleared for the option."
\\[5pt] - Trigger: "change of plans ... request transition to Palo Alto"
\\ ~~Response must be:
\\ ~~"Cessna 715 Mike Tango, transition approved, proceed toward Palo Alto."
\\ ~~(Do not say contact ground 125.0 here.)
\\[5pt] - Trigger: "We will follow 880 freeway ..."
\\ ~~Response must be:
\\ ~~"Roger."
\\[5pt] H) Palo Alto arrival segment
\\ - Trigger: "Palo Alto Tower ... 8 miles north ... inbound for the option ..."
\\ ~~Response must be pattern entry:
\\ ~~"Cessna 715 Mike Tango, enter right traffic runway 31, report midfield downwind."
\\[5pt] - Trigger: pilot readback "Runway 31 right traffic ..."
\\ ~~Response must be:
\\ ~~"Roger."
\\[5pt] - Trigger: "Midfield right downwind runway 31 ..."
\\ ~~Response must be:
\\ ~~"Cessna 715 Mike Tango, runway 31 cleared for the option."
\\[5pt] - Trigger: pilot says "full stop"
\\ ~~Response must be:
\\ ~~"Roger, full stop."
\\[5pt] - Trigger: pilot readback "Taxi via Zulu, Yankee 2, monitor ground ..."
\\ ~~Response must be:
\\ ~~"Roger."
\\ ~~(Do NOT re-issue taxi instructions.)
\\[5pt] If the pilot checks in with San Francisco Tower (at 1400) and then complies with "Ident", the next controller transmission MUST issue the Class Bravo clearance: "Cessna 715 Mike Tango, cleared through the Class Bravo, maintain VFR at or below 2000, keep the highway off your right."
\\[5pt] Now respond to the next pilot transmission.}
\end{tcolorbox}
\begin{tcolorbox}[breakable, enhanced, colback=gray!7, colframe=gray!60!black, colbacktitle=gray!60!black, boxrule=0.8pt, fonttitle=\bfseries, coltitle=white, title={ICAO phraseology style block (appended to every prompt P1--P5)}]
{\ttfamily\footnotesize\raggedright PHRASEOLOGY STYLE (follow exactly):
\\ - Use the abbreviated callsign (the last three characters) in every transmission,
\\ ~~including the first contact: "Cessna 715 Mike Tango" becomes "5 Mike Tango".
\\ - Read altimeter settings, headings, frequencies and squawk codes digit by digit,
\\ ~~saying "niner" for 9 and "decimal" for the point. Do NOT digit-spell altitudes.
\\ - One short transmission. Output the controller's words only (no labels). Use a brief
\\ ~~"good day" sign-off when handing the aircraft to another frequency.}
\end{tcolorbox}

\clearpage
\section{Full Flight Transcripts}
\label{app:transcripts}
The two boxes below give the complete hand-transcribed radio exchanges for both flights.
The first (violet) is the evaluation scenario P0 against which every model is scored; the
second (teal) is the disjoint flight used only as the in-context example. Each turn is
tagged with a \colorbox{pilotbadge}{\textcolor{white}{\scriptsize\bfseries\,PILOT\,}}
or \colorbox{atcbadge}{\textcolor{white}{\scriptsize\bfseries\,\ ATC\ \,}} badge.

\definecolor{pilotbadge}{HTML}{2A6FB0}
\definecolor{atcbadge}{HTML}{C0603A}
\definecolor{flightmain}{HTML}{6A4C93}
\definecolor{flighticl}{HTML}{2A8A7F}
\providecommand{\faPlane}{}
\providecommand{\faBroadcastTower}{}
\providecommand{\pilotsay}[1]{\par\smallskip\noindent\colorbox{pilotbadge}{\textcolor{white}{\scriptsize\bfseries\,\faPlane\ PILOT\,}}\hspace{4pt}#1}
\providecommand{\atcsay}[1]{\par\smallskip\noindent\colorbox{atcbadge}{\textcolor{white}{\scriptsize\bfseries\,\faBroadcastTower\ ATC\,}}\hspace{4pt}#1}
\tcbset{before skip=6pt, after skip=6pt}

\begin{tcolorbox}[breakable, enhanced, colback=flightmain!6, colframe=flightmain, colbacktitle=flightmain, boxrule=0.9pt, fonttitle=\bfseries, coltitle=white, title={Evaluation flight P0 --- San Francisco ``Bay Tour,'' Cessna 715 Mike Tango (departing KPAO)}]
\begin{center}\includegraphics[width=0.62\linewidth]{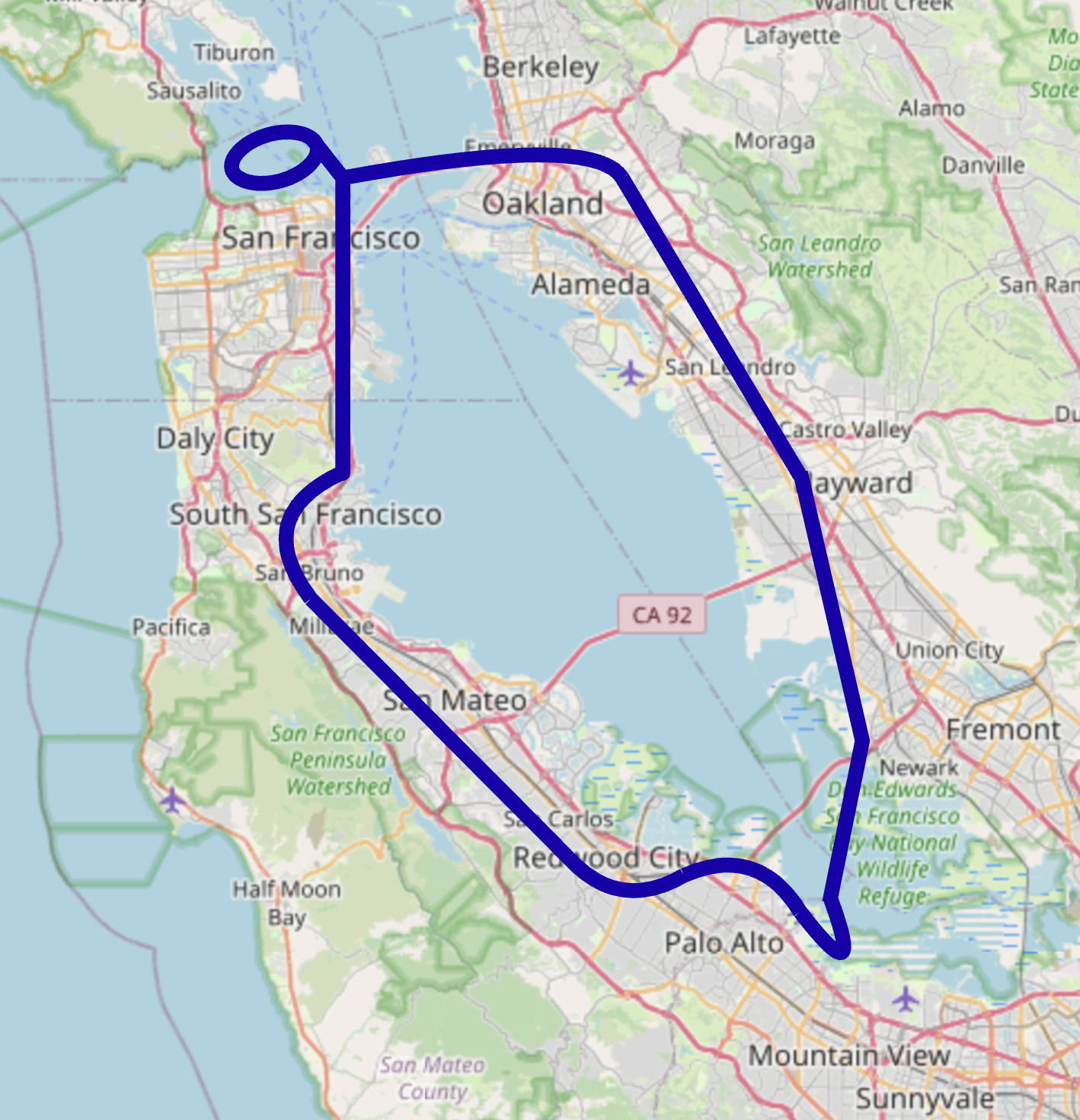}\end{center}
\vspace{2pt}\footnotesize
\pilotsay{Palo Alto Ground Cessna 715 Mike Tango run up complete}
\atcsay{5 Mike Tango hold short of Yankee 1 and contact tower good day}
\pilotsay{Palo Alto Tower Cessna 715 Mike Tango holding short of Yankee 1}
\atcsay{Palo alto tower information victor now current wind 330 at one five altimeter two niner niner two runway 31 in use use caution for increase bird activity full length over runway Cessna 715 Mike Tango Palo Alto tower runway 31 cleared for takeoff wind 330 at one five}
\pilotsay{Runway 31 cleared for takeoff 715 Mike Tango}
\atcsay{Cessna 5 Mike Tango contact San Carlos tower at 119.0}
\pilotsay{San Carlos on 119.0 have a good day 715 Mike Tango}
\pilotsay{San Carlos Tower Cessna 715 Mike Tango at 600}
\atcsay{715 Mike Tango San Carlos Tower intercept and follow 101 keep it off to your right during your transition maintain VFR conditions at or below 1500 San Carlos altimeter is two niner niner two}
\pilotsay{VFR at or below 1500 will keep the highway off our right 715 Mike Tango}
\atcsay{Cessna 715 Mike Tango, remain outside of Bravo to receive authorization contact San Francisco tower 120.5}
\pilotsay{San Francisco Tower at 120.5 have a good day 715 Mike Tango}
\pilotsay{San Francisco Tower Cessna 715 Mike Tango at 1400}
\atcsay{Cessna 715 Mike Tango San Francisco tower ident}
\pilotsay{Ident}
\atcsay{Cessna 5 Mike Tango radar contact about northwest San Carlos airport, clears through the Bravo airspace keep the 101 off the right side maintain VFR at or below 2000 while in the Bravo. San francisco altimeter two niner niner three, say altitude}
\pilotsay{Cleared through the Class Bravo, keep the highway off our right, at or below 2000, 715 Mike Tango, altitude 1500}
\atcsay{Thank you 715 Mike Tango traffic departing San Francisco westbound I have the airbus [ ] report the traffic in sight 715 Mike Tango did you say you have that traffic on the runway departing westbound in sight}
\pilotsay{Affirmative traffic in sight 715 Mike Tango}
\atcsay{715 Mike Tango pass behind that traffic caution wake turbulence and resume your transition if you want to do a 360 there that is approved to get the space}
\pilotsay{We'll do a 360 715 Mike Tango}
\atcsay{715 Mike Tango leaving Bravo [ ] contact approach at 135.1}
\pilotsay{NorCal Approach on 135.1 715 Mike Tango}
\pilotsay{NorCal Approach Cessna 715 Mike Tango at 1900}
\atcsay{Cessna 715 Mike Tango norcal depart traffic Cessna 715 Mike Tango you have exited class Bravo airspace altitude your discretion remain north and west of the bay bridge}
\pilotsay{Altitude at our discretion we will stay west of the Bay Bridge 715 Mike Tango}
\pilotsay{NorCal Approach Cessna 715 Mike Tango requesting Oakland Charlie transition, information Hotel, to Hayward}
\atcsay{Cessna 5 Mike Tango cross north of the bay bridge [ ] and then over the Coliseum at 2500}
\pilotsay{Cross north of the Bay Bridge then 2500, 715 Mike Tango}
\atcsay{Cessna 5 Mike Tango traffic one o clock one mile currently north westbound in-bound for the bay tour 1800 Cessna}
\pilotsay{Looking for traffic 715 Mike Tango}
\pilotsay{Traffic in sight 715 Mike Tango}
\atcsay{November 5 Mike Tango roger contact oakland tower 118.3}
\pilotsay{Oakland 118.3 have a good evening 715 Mike Tango}
\pilotsay{Oakland Tower Cessna 715 Mike Tango at 2500 transitioning to Hayward}
\atcsay{November 715 Mike Tango oakland tower follow the 880 freeway southeast bound the oakland altimeter is two niner niner five and for now maintain VFR at or above 1500 for traffic}
\pilotsay{We will follow the highway at or above 1500 715 Mike Tango}
\atcsay{715 Mike Tango traffic twelve o clock in one mile southwest bound 600 helicopter}
\pilotsay{Looking for traffic 715 Mike Tango}
\atcsay{November 5 Mike Tango traffic no factor VFR decent to at or below 1400 at your discretion}
\pilotsay{We will stay at or below 1400 715 Mike Tango}
\atcsay{November 5 Mike Tango radar services terminated keep the code contact Hayward tower have a good day}
\pilotsay{We will keep code and contact Hayward 715 Mike Tango}
\pilotsay{Hayward Tower Cessna 715 Mike Tango at 1600 requesting inbound for the option, information Hotel, request runway 28 Left}
\atcsay{Cessna 715 Mike Tango Hayward tower again I would not recommend any pattern work at the moment there are multiple flocks of birds on both runways}
\pilotsay{Hayward Tower Cessna 715 Mike Tango change of plans, thanks for the bird warning, request transition to Palo Alto}
\atcsay{Cessna 715 Mike Tango roger proceed as requested}
\pilotsay{We will follow 880 freeway 715 Mike Tango}
\atcsay{Cessna 5 Mike Tango you can proceed direct towards palo alto there's no traffic on final in oakland Cessna 5 Mike Tango contact palo alto tower}
\pilotsay{Palo Alto Tower Cessna 715 Mike Tango 8 miles north, inbound for the option, information Victor}
\atcsay{Cessna 715 Mike Tango Palo alto tower make right traffic runway 31 traffic at your ten o'clock in about four miles southwest bound altitude indicating 1500 type unknown if able maintain at or below 1000}
\pilotsay{Runway 31 right traffic, will watch for the traffic 715 Mike Tango}
\atcsay{Cessna 5 Mike Tango in order to pass behind that traffic turn left 15 degrees that traffic is now twelve o clock two miles still southwest bound}
\pilotsay{Traffic in sight, turning left 715 Mike Tango}
\atcsay{Roger passing trail of that aircraft enter midfield right downwind runway 31}
\pilotsay{Midfield right downwind runway 31 715 Mike Tango}
\atcsay{Cessna 715 Mike Tango you will be number three following a red citabria ahead of you at your twelve o clock two miles ahead of you setting you at a 1500 number three runway 31 cleared for the option}
\pilotsay{Runway 31 cleared for the option number three 715 Mike Tango}
\pilotsay{Palo Alto Tower Cessna 715 Mike Tango full stop}
\atcsay{Cessna 715 mike tango roger Mike tango traffic to follow is now on the right base turn altitude indicating 500}
\pilotsay{Following traffic 715 Mike Tango}
\atcsay{Cessna 5 Mike Tango taxi parking zulu yankee 2 and monitor ground good night}
\pilotsay{Taxi via Zulu, Yankee 2, monitor ground, 715 Mike Tango}
\end{tcolorbox}

\vspace{4pt}
\begin{tcolorbox}[breakable, enhanced, colback=flighticl!6, colframe=flighticl, colbacktitle=flighticl, boxrule=0.9pt, fonttitle=\bfseries, coltitle=white, title={In-context example flight --- Reid--Hillview, Cirrus 485 Echo Mike Romeo}]
\begin{center}\includegraphics[width=0.62\linewidth]{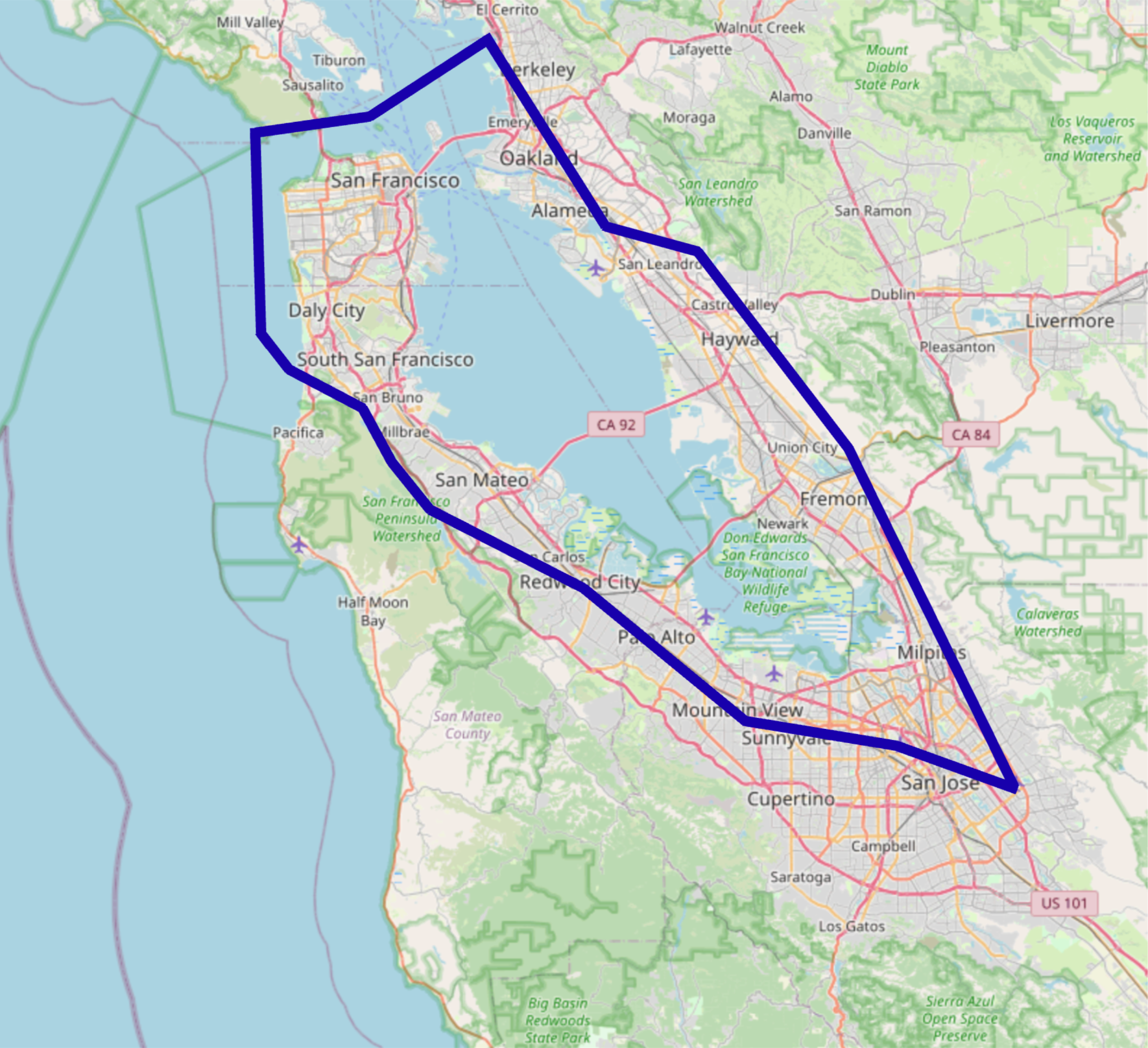}\end{center}
\vspace{2pt}\footnotesize
\pilotsay{Reid Hill View Ground Cirrus 485 Echo Mike Romeo Hangers taxi to the run up, and this will be a bay tour clockwise, san jose transition we have information yankee flight following}
\atcsay{Cirrus 485 Echo Mike Reid hill view ground, taxi to the run-up via zulu}
\pilotsay{To the run-up via zulu, 5 echo mike}
\pilotsay{And Ground Cirrus 485 echo mike, run-up complete}
\atcsay{Cirrus 5 echo mike, 31 right taxi to the runway via zulu, zulu 2 yankee 2}
\pilotsay{31 right taxi via zulu, zulu 2, yankee 2}
\pilotsay{Reid Hill View Tower Cirrus 485 Echo Mike, holding short of 31 right at zulu 2, ready for departure left crosswind, bay tour, san jose airspace}
\atcsay{Cirrus 485 echo mike reid hill view tower, fly straight out, 31 right clear for takeoff}
\pilotsay{Straight out 31 right, clear for takeoff 485 echo mike}
\atcsay{Cirrus 485 Echo mike, squawk 0330}
\pilotsay{0330, 5 echo mike}
\atcsay{Cirrus 5 Echo mike, contact san jose tower at 124.0}
\pilotsay{124.0, 5 echo mike}
\pilotsay{San Jose Tower Cirrus 485 echo mike, out of reid hill view 800 climbing 2 thousand 800, requesting transition through your airspace for a bay tour}
\atcsay{Cirrus 485 Echo mike, San Jose tower ident, san jose altimeter two niner niner four, cross san jose midfield two thousand}
\pilotsay{two niner niner four, cross san jose midfield two thousand, 5 echo mike}
\atcsay{Cirrus 5 Echo mike radar contact 4 miles east of san jose}
\pilotsay{position checks, 5 echo mike}
\atcsay{Cirrus 5 Echo mike fly heading 270 at or below 2500}
\pilotsay{270 at or below 2500, 5 echo mike}
\atcsay{Cirrus 5 Echo mike contact norcal approach 120.1}
\pilotsay{120.1, 5 echo mike}
\pilotsay{Norcal Cirrus 485 echo mike checking in on a 270 heading at or below 2500 bay tour}
\atcsay{485 echo mike norcal departure, resume navigation, remain southwest of the bay shore freeway}
\pilotsay{southwest of the bay shore freeway, resume on nav, 5 echo mike}
\atcsay{5 echo mike contact approach 135.65}
\pilotsay{135.65, 5 echo mike}
\pilotsay{norcal cirrus 485 echo mike, level 2500 on a bay tour}
\atcsay{485 echo mike san francisco altimeter two niner niner five}
\pilotsay{two niner niner five, 5 echo mike}
\pilotsay{norcal cirrus 485 echo mike with a question}
\atcsay{485 echo mike go}
\pilotsay{cirrus 485 echo mike is there an active TFR over san francisco, and will we be able to do the transition, or will we have to do the pacifica one}
\atcsay{5 echo mike tower is not accepting transitions right now so whatever is outside of that you can plan to do that}
\pilotsay{roger thanks, 5 echo mike}
\pilotsay{norcal cirrus 485 echo mike we are going to climb all the way up to 3700, but we'll stay below bravo and go north up shore side.}
\atcsay{485 echo mike, roger, understand your going to be going pacifica half moon bay and that coast?}
\pilotsay{Thats correct, I asked the question because we see a thin layer of fog and we think we'll be ok at 3700}
\atcsay{5 echo mike thanks Cirrus 5 echo mike traffic 12 o-clock 5 miles east bound a 737 at 8000 descending 5000}
\pilotsay{Negative contact, we're looking, 5 echo mike}
\pilotsay{Cirrus 5 echo mike, we have the traffic in sight, and we are going to stay at 3500 for now}
\atcsay{5 echo mike roger 5 echo mike contact approach at 135.1}
\pilotsay{135.1, 5 echo mike, good day}
\pilotsay{norcal cirrus 485 echo mike, level 3500, we'll be requesting the pacifica transition}
\atcsay{Cirrus 485 echo mike norcal approach mike san francisco altimeter two niner niner five, remain outside of bravo just up the coastline}
\pilotsay{Ok outside of bravo, two niner niner five, we'll let you know, 5 echo mike}
\pilotsay{And norcal cirrus 485 echo mike with an update we're going to stay at 2900 and we'll stay clear of the bravo and do the transition outside of the bravo}
\atcsay{5 echo mike thank you sir 5 echo mike traffic eleven oclock at about 2 miles south bound at a thousand feet a cessna}
\pilotsay{Negative contact we are looking, 5 echo mike}
\pilotsay{Norcal approach Cirrus 485 echo mike another quick update we are going to stay on this heading for another 3 miles then we will turn right towards mormon temple, then reid hill view}
\atcsay{November 5 echo mike roger reid hill view is your destination, I'll update that November 5 echo mike information zulu is current, reid hill view altimeter two niner niner three}
\pilotsay{two niner niner three, we'll let you know about zulu, 5 echo mike}
\atcsay{November 5 echo mike Oakland tower 118.3 good day}
\pilotsay{118.3 good day, 5 echo mike}
\pilotsay{Oakland Tower cirrus 485 echo mike two thousand niner hundred level, we are going to reid hill view}
\atcsay{November 485 echo mike Oakland tower altimeter 2997, exit the class charlie airspace north of lake chabot}
\pilotsay{2997, exit the class charlie airspace north of lake chabot, 5 echo mike}
\atcsay{November 5 echo mike maintain VFR at or below 2500}
\pilotsay{VFR at or below 2500, 5 echo mike}
\atcsay{November 5 echo mike contact norcal 125.35}
\pilotsay{125.35, good day 5 echo mike}
\pilotsay{Norcal Cirrus 485 echo make level 2500 going to reid hill view, we don't have the ATIS yet}
\atcsay{November 485 echo mike norcal approach VFR altitude at your discretion, resume on navigation reid hill view Oakland altimeter 2997}
\pilotsay{2997 own nav, 5 echo mike}
\atcsay{November 5 echo mike traffic 12 oclock in 2 miles northwest bound a cessna}
\pilotsay{Negative contact turning 20 to the left, 5 echo mike}
\pilotsay{Cirrus 485 echo mike, traffic in sight, no factor}
\pilotsay{Norcal Cirrus 485 echo mike level 2500 going to reid hill view with zulu}
\atcsay{November 485 echo mike norcal approach, san jose altimeter 2994}
\pilotsay{2994, 5 echo mike}
\pilotsay{Norcal Cirrus 485 echo mike we have reid hill view in sight}
\atcsay{5 echo mike roger traffic 1 o'clock 3 miles a banner toe, contact reid hill view tower, 119.8}
\pilotsay{Over to reid hill view tower 119.8, and we'll keep an eye on the traffic}
\pilotsay{Reid hill view tower Cirrus 485 echo mike 2500, inbound with zulu, fullstop}
\atcsay{Cirrus 485 echo mike reid hill view tower, enter right traffic runway 31 right}
\pilotsay{Enter right traffic runway 31 right, 5 echo mike}
\atcsay{Cirrus 485 echo mike runway 31 right cleared to land}
\pilotsay{Cleared to land runway 31 right, 5 echo mike}
\atcsay{Cirrus 5 echo mike if able turn right at delta and contact ground}
\pilotsay{Right at delta, contact ground, 5 echo mike}
\end{tcolorbox}

\vspace{4pt}

\clearpage
\section{LLM-as-Judge Rubric}
\label{app:judge}
The judge (GPT-5.5) receives the preceding exchange, the current pilot transmission, a
single reference controller reply, and the candidate output, and returns integer scores
on the seven dimensions below plus a short free-text rationale. The reference is
provided as \emph{one} correct example, not a target to match, so that valid
alternative phrasings are not penalised.

\begin{tcolorbox}[breakable, enhanced, colback=violet!7, colframe=violet!60!black, colbacktitle=violet!60!black, boxrule=0.8pt, fonttitle=\bfseries, coltitle=white, title={LLM-as-judge system prompt (GPT-5.5, low reasoning effort)}]
{\ttfamily\footnotesize\raggedright You are a senior FAA air traffic controller acting as an expert evaluator. You will see
\\ a short slice of a pilot-ATC radio exchange, the current pilot transmission, and a
\\ CANDIDATE controller reply produced by an AI. A REFERENCE reply (one correct example)
\\ is provided; the candidate need NOT match it verbatim, as multiple phrasings can be valid.
\\[5pt] Score the CANDIDATE from 1 (poor) to 5 (excellent) on each dimension:
\\ - phraseology: standard FAA/ICAO phraseology and format
\\ - correctness: operationally correct and appropriate response to this pilot call
\\ - callsign: correct and properly abbreviated aircraft callsign usage
\\ - conciseness: appropriately brief; no extraneous content
\\ - safety: contains no unsafe or incorrect instruction
\\ - hallucination\_free: invents no frequencies/weather/altimeter/traffic not given
\\ - overall: overall quality as an ATC transmission
\\[5pt] Respond ONLY as compact JSON with integer scores and a rationale of at most 20 words.}
\end{tcolorbox}

\section{Qualitative Examples}
\label{app:qual}
The five boxes below are drawn from the human-annotated subset, so each carries both
LLM-judge and human scores on all seven dimensions. The first three isolate short-history
turns and span the range of agreement: at turn~8 the judge and human both rate a clean
check-in 5/5; at turn~23 the judge scores the transmission 1/5 while the human gives 5/5,
the leniency gap in miniature; at turn~34 an airborne aircraft is answered with a ground
taxi instruction, and both raters mark it down. The last two boxes show \emph{long-history}
turns: at the same late turn~34, conditioned on the full 33-turn conversation,
GPT-5.4-mini keeps the thread and clears the aircraft correctly (5/5), whereas
Qwen2.5-14B loses it, issuing a premature taxi instruction to a still-landing aircraft
(judge 1/5). Each box shows the dialogue history, the pilot transmission, the ground-truth
reply (blue), the model output (orange), the automatic similarity scores, the judge's
rationale, and the side-by-side judge/human score table.

\tcbset{before skip=6pt, after skip=6pt}
\begin{tcolorbox}[breakable, enhanced, colback=green!7, colframe=green!60!black, colbacktitle=green!60!black, boxrule=0.9pt, fonttitle=\bfseries, coltitle=white, title={Example 1\quad$\vert$\quad Turn 8\quad$\vert$\quad Claude-Sonnet-4.6\quad$\vert$\quad short history: judge and human agree}]
{\scriptsize\itshape\color{black!60}Pilot: VFR at or below 1500 will keep the highway off our right 715 Mike Tango\\ ATC: Cessna 715 Mike Tango, remain outside of Bravo to receive authorization contact San Francisco tower 120.5\\ Pilot: San Francisco Tower at 120.5 have a good day 715 Mike Tango}\par\smallskip
\textbf{Pilot.}\; San Francisco Tower Cessna 715 Mike Tango at 1400\par\smallskip
\begin{tcolorbox}[enhanced, boxrule=0.6pt, arc=1mm, left=2mm, right=2mm, top=1.2mm, bottom=1.2mm, colback=blue!8, colframe=blue!65!black]\textcolor{blue!65!black}{\textbf{Ground-truth ATC}}\; Cessna 715 Mike Tango San Francisco tower ident\end{tcolorbox}
\vspace{2pt}
\begin{tcolorbox}[enhanced, boxrule=0.6pt, arc=1mm, left=2mm, right=2mm, top=1.2mm, bottom=1.2mm, colback=orange!8, colframe=orange!65!black]\textcolor{orange!65!black}{\textbf{Model output}}\; Cessna 715 Mike Tango San Francisco Tower, ident\end{tcolorbox}
\smallskip\par
\noindent\begin{minipage}[t]{0.40\linewidth}
{\footnotesize\begin{tabular}{@{}lcc@{}}\toprule
Dimension & \textcolor{red!60!black}{Judge} & \textcolor{teal!65!black}{Human} \\
\midrule
Phraseology & 5 & 5 \\
Correctness & 5 & 5 \\
Callsign & 5 & 5 \\
Conciseness & 5 & 5 \\
Safety & 5 & 5 \\
Halluc.-free & 5 & 5 \\
\textbf{Overall} & \textbf{5} & \textbf{5} \\
\bottomrule\end{tabular}}
\end{minipage}\hfill\begin{minipage}[t]{0.56\linewidth}
\footnotesize\vspace{0pt}
\textbf{Lexical (vs.\ ground truth):}\\ ROUGE-L~1.00\quad BLEU-1~0.88\quad Semantic~0.99\par\smallskip
\textbf{Judge rationale:}\\ \textit{Matches appropriate tower response; standard ident instruction with correct callsign and no invented information.}
\end{minipage}
\end{tcolorbox}

\begin{tcolorbox}[breakable, enhanced, colback=orange!7, colframe=orange!60!black, colbacktitle=orange!60!black, boxrule=0.9pt, fonttitle=\bfseries, coltitle=white, title={Example 2\quad$\vert$\quad Turn 23\quad$\vert$\quad Gemma-2-9B\quad$\vert$\quad short history: judge harsh, human lenient}]
{\scriptsize\itshape\color{black!60}ATC: November 715 Mike Tango oakland tower follow the 880 freeway southeast bound the oakland altimeter is two niner niner five and for now maintain VFR at or above 1500 for traffic\\ Pilot: We will follow the highway at or above 1500 715 Mike Tango\\ ATC: 715 Mike Tango traffic twelve o clock in one mile southwest bound 600 helicopter}\par\smallskip
\textbf{Pilot.}\; Looking for traffic 715 Mike Tango\par\smallskip
\begin{tcolorbox}[enhanced, boxrule=0.6pt, arc=1mm, left=2mm, right=2mm, top=1.2mm, bottom=1.2mm, colback=blue!8, colframe=blue!65!black]\textcolor{blue!65!black}{\textbf{Ground-truth ATC}}\; November 5 Mike Tango traffic no factor VFR decent to at or below 1400 at your discretion\end{tcolorbox}
\vspace{2pt}
\begin{tcolorbox}[enhanced, boxrule=0.6pt, arc=1mm, left=2mm, right=2mm, top=1.2mm, bottom=1.2mm, colback=orange!8, colframe=orange!65!black]\textcolor{orange!65!black}{\textbf{Model output}}\; 5 Mike Tango, maintain VFR at or below 1500, remain clear of controlled airspace.  Traffic inbound runway 31 left.\end{tcolorbox}
\smallskip\par
\noindent\begin{minipage}[t]{0.40\linewidth}
{\footnotesize\begin{tabular}{@{}lcc@{}}\toprule
Dimension & \textcolor{red!60!black}{Judge} & \textcolor{teal!65!black}{Human} \\
\midrule
Phraseology & 2 & 5 \\
Correctness & 1 & 5 \\
Callsign & 4 & 5 \\
Conciseness & 3 & 5 \\
Safety & 2 & 5 \\
Halluc.-free & 1 & 5 \\
\textbf{Overall} & \textbf{1} & \textbf{5} \\
\bottomrule\end{tabular}}
\end{minipage}\hfill\begin{minipage}[t]{0.56\linewidth}
\footnotesize\vspace{0pt}
\textbf{Lexical (vs.\ ground truth):}\\ ROUGE-L~0.39\quad BLEU-1~0.37\quad Semantic~0.63\par\smallskip
\textbf{Judge rationale:}\\ \textit{Invents runway traffic, gives inappropriate airspace instruction, and changes altitude contrary to context without resolving reported traffic.}
\end{minipage}
\end{tcolorbox}

\begin{tcolorbox}[breakable, enhanced, colback=red!7, colframe=red!60!black, colbacktitle=red!60!black, boxrule=0.9pt, fonttitle=\bfseries, coltitle=white, title={Example 3\quad$\vert$\quad Turn 34\quad$\vert$\quad Gemma-2-9B\quad$\vert$\quad short history: unsafe output, both raters mark it down}]
{\scriptsize\itshape\color{black!60}Pilot: Midfield right downwind runway 31 715 Mike Tango\\ ATC: Cessna 715 Mike Tango you will be number three following a red citabria ahead of you at your twelve o clock two miles ahead of you setting you at a 1500 number three runway 31 cleared for the option\\ Pilot: Runway 31 cleared for the option number three 715 Mike Tango}\par\smallskip
\textbf{Pilot.}\; Palo Alto Tower Cessna 715 Mike Tango full stop\par\smallskip
\begin{tcolorbox}[enhanced, boxrule=0.6pt, arc=1mm, left=2mm, right=2mm, top=1.2mm, bottom=1.2mm, colback=blue!8, colframe=blue!65!black]\textcolor{blue!65!black}{\textbf{Ground-truth ATC}}\; Cessna 715 mike tango roger
Mike tango traffic to follow is now on the right base turn altitude indicating 500\end{tcolorbox}
\vspace{2pt}
\begin{tcolorbox}[enhanced, boxrule=0.6pt, arc=1mm, left=2mm, right=2mm, top=1.2mm, bottom=1.2mm, colback=orange!8, colframe=orange!65!black]\textcolor{orange!65!black}{\textbf{Model output}}\; Cessna 715 Mike Tango, taxi to the run-up via zulu.\end{tcolorbox}
\smallskip\par
\noindent\begin{minipage}[t]{0.40\linewidth}
{\footnotesize\begin{tabular}{@{}lcc@{}}\toprule
Dimension & \textcolor{red!60!black}{Judge} & \textcolor{teal!65!black}{Human} \\
\midrule
Phraseology & 1 & 5 \\
Correctness & 1 & 1 \\
Callsign & 5 & 5 \\
Conciseness & 4 & 3 \\
Safety & 1 & 1 \\
Halluc.-free & 1 & 1 \\
\textbf{Overall} & \textbf{1} & \textbf{2} \\
\bottomrule\end{tabular}}
\end{minipage}\hfill\begin{minipage}[t]{0.56\linewidth}
\footnotesize\vspace{0pt}
\textbf{Lexical (vs.\ ground truth):}\\ ROUGE-L~0.39\quad BLEU-1~0.18\quad Semantic~0.56\par\smallskip
\textbf{Judge rationale:}\\ \textit{Incorrectly issues ground taxi/run-up instruction to airborne aircraft; invents route and creates unsafe confusion.}
\end{minipage}
\end{tcolorbox}

\begin{tcolorbox}[breakable, enhanced, colback=blue!7, colframe=blue!60!black, colbacktitle=blue!60!black, boxrule=0.9pt, fonttitle=\bfseries, coltitle=white, title={Example 4\quad$\vert$\quad Turn 34\quad$\vert$\quad GPT-5.4-mini\quad$\vert$\quad long history: model keeps the thread over 33 turns}]
{\scriptsize\itshape\color{black!60}\textit{\color{black!45}(turns 1--23 omitted)}\\ \textbf{P:} We will stay at or below 1400 715 Mike Tango\\ \textbf{A:} November 5 Mike Tango radar services terminated keep the code contact Hayward tower have a good day\\ \textbf{P:} We will keep code and contact Hayward 715 Mike Tango\\ \textbf{P:} Hayward Tower Cessna 715 Mike Tango at 1600 requesting inbound for the option, information Hotel, request runway 28 Left\\ \textbf{A:} Cessna 715 Mike Tango Hayward tower again I would not recommend any pattern work at the moment there are multiple flocks of birds on both runways\\ \textbf{P:} Hayward Tower Cessna 715 Mike Tango change of plans, thanks for the bird warning, request transition to Palo Alto\\ \textbf{A:} Cessna 715 Mike Tango roger proceed as requested\\ \textbf{P:} We will follow 880 freeway 715 Mike Tango\\ \textbf{A:} Cessna 5 Mike Tango you can proceed direct towards palo alto there's no traffic on final in oakland Cessna 5 Mike Tango contact palo alto tower\\ \textbf{P:} Palo Alto Tower Cessna 715 Mike Tango 8 miles north, inbound for the option, information Victor\\ \textbf{A:} Cessna 715 Mike Tango Palo alto tower make right traffic runway 31 traffic at your ten o'clock in about four miles southwest bound altitude indicating 1500 type unknown if able maintain at or below 1000\\ \textbf{P:} Runway 31 right traffic, will watch for the traffic 715 Mike Tango\\ \textbf{A:} Cessna 5 Mike Tango in order to pass behind that traffic turn left 15 degrees that traffic is now twelve o clock two miles still southwest bound\\ \textbf{P:} Traffic in sight, turning left 715 Mike Tango\\ \textbf{A:} Roger passing trail of that aircraft enter midfield right downwind runway 31\\ \textbf{P:} Midfield right downwind runway 31 715 Mike Tango\\ \textbf{A:} Cessna 715 Mike Tango you will be number three following a red citabria ahead of you at your twelve o clock two miles ahead of you setting you at a 1500 number three runway 31 cleared for the option\\ \textbf{P:} Runway 31 cleared for the option number three 715 Mike Tango}\par\smallskip
\textbf{Pilot.}\; Palo Alto Tower Cessna 715 Mike Tango full stop\par\smallskip
\begin{tcolorbox}[enhanced, boxrule=0.6pt, arc=1mm, left=2mm, right=2mm, top=1.2mm, bottom=1.2mm, colback=blue!8, colframe=blue!65!black]\textcolor{blue!65!black}{\textbf{Ground-truth ATC}}\; Cessna 715 mike tango roger
Mike tango traffic to follow is now on the right base turn altitude indicating 500\end{tcolorbox}
\vspace{2pt}
\begin{tcolorbox}[enhanced, boxrule=0.6pt, arc=1mm, left=2mm, right=2mm, top=1.2mm, bottom=1.2mm, colback=orange!8, colframe=orange!65!black]\textcolor{orange!65!black}{\textbf{Model output}}\; Cessna 715 Mike Tango runway 31 cleared to land.\end{tcolorbox}
\smallskip\par
\noindent\begin{minipage}[t]{0.40\linewidth}
{\footnotesize\begin{tabular}{@{}lcc@{}}\toprule
Dimension & \textcolor{red!60!black}{Judge} & \textcolor{teal!65!black}{Human} \\
\midrule
Phraseology & 5 & 5 \\
Correctness & 5 & 5 \\
Callsign & 5 & 5 \\
Conciseness & 5 & 5 \\
Safety & 5 & 5 \\
Halluc.-free & 5 & 5 \\
\textbf{Overall} & \textbf{5} & \textbf{5} \\
\bottomrule\end{tabular}}
\end{minipage}\hfill\begin{minipage}[t]{0.56\linewidth}
\footnotesize\vspace{0pt}
\textbf{Lexical (vs.\ ground truth):}\\ ROUGE-L~0.34\quad BLEU-1~0.16\quad Semantic~0.53\par\smallskip
\textbf{Judge rationale:}\\ \textit{Standard, concise landing clearance using correct callsign and runway; no invented or unsafe information.}
\end{minipage}
\end{tcolorbox}

\begin{tcolorbox}[breakable, enhanced, colback=red!7, colframe=red!60!black, colbacktitle=red!60!black, boxrule=0.9pt, fonttitle=\bfseries, coltitle=white, title={Example 5\quad$\vert$\quad Turn 34\quad$\vert$\quad Qwen2.5-14B\quad$\vert$\quad long history: model loses the thread late in the flight}]
{\scriptsize\itshape\color{black!60}\textit{\color{black!45}(turns 1--23 omitted)}\\ \textbf{P:} We will stay at or below 1400 715 Mike Tango\\ \textbf{A:} November 5 Mike Tango radar services terminated keep the code contact Hayward tower have a good day\\ \textbf{P:} We will keep code and contact Hayward 715 Mike Tango\\ \textbf{P:} Hayward Tower Cessna 715 Mike Tango at 1600 requesting inbound for the option, information Hotel, request runway 28 Left\\ \textbf{A:} Cessna 715 Mike Tango Hayward tower again I would not recommend any pattern work at the moment there are multiple flocks of birds on both runways\\ \textbf{P:} Hayward Tower Cessna 715 Mike Tango change of plans, thanks for the bird warning, request transition to Palo Alto\\ \textbf{A:} Cessna 715 Mike Tango roger proceed as requested\\ \textbf{P:} We will follow 880 freeway 715 Mike Tango\\ \textbf{A:} Cessna 5 Mike Tango you can proceed direct towards palo alto there's no traffic on final in oakland Cessna 5 Mike Tango contact palo alto tower\\ \textbf{P:} Palo Alto Tower Cessna 715 Mike Tango 8 miles north, inbound for the option, information Victor\\ \textbf{A:} Cessna 715 Mike Tango Palo alto tower make right traffic runway 31 traffic at your ten o'clock in about four miles southwest bound altitude indicating 1500 type unknown if able maintain at or below 1000\\ \textbf{P:} Runway 31 right traffic, will watch for the traffic 715 Mike Tango\\ \textbf{A:} Cessna 5 Mike Tango in order to pass behind that traffic turn left 15 degrees that traffic is now twelve o clock two miles still southwest bound\\ \textbf{P:} Traffic in sight, turning left 715 Mike Tango\\ \textbf{A:} Roger passing trail of that aircraft enter midfield right downwind runway 31\\ \textbf{P:} Midfield right downwind runway 31 715 Mike Tango\\ \textbf{A:} Cessna 715 Mike Tango you will be number three following a red citabria ahead of you at your twelve o clock two miles ahead of you setting you at a 1500 number three runway 31 cleared for the option\\ \textbf{P:} Runway 31 cleared for the option number three 715 Mike Tango}\par\smallskip
\textbf{Pilot.}\; Palo Alto Tower Cessna 715 Mike Tango full stop\par\smallskip
\begin{tcolorbox}[enhanced, boxrule=0.6pt, arc=1mm, left=2mm, right=2mm, top=1.2mm, bottom=1.2mm, colback=blue!8, colframe=blue!65!black]\textcolor{blue!65!black}{\textbf{Ground-truth ATC}}\; Cessna 715 mike tango roger
Mike tango traffic to follow is now on the right base turn altitude indicating 500\end{tcolorbox}
\vspace{2pt}
\begin{tcolorbox}[enhanced, boxrule=0.6pt, arc=1mm, left=2mm, right=2mm, top=1.2mm, bottom=1.2mm, colback=orange!8, colframe=orange!65!black]\textcolor{orange!65!black}{\textbf{Model output}}\; Cessna 715 Mike Tango your frequency is now closed contact ground on 121.2 for taxi and tie down instructions good day\end{tcolorbox}
\smallskip\par
\noindent\begin{minipage}[t]{0.40\linewidth}
{\footnotesize\begin{tabular}{@{}lcc@{}}\toprule
Dimension & \textcolor{red!60!black}{Judge} & \textcolor{teal!65!black}{Human} \\
\midrule
Phraseology & 1 & 1 \\
Correctness & 1 & 1 \\
Callsign & 5 & 5 \\
Conciseness & 2 & 1 \\
Safety & 1 & 1 \\
Halluc.-free & 1 & 1 \\
\textbf{Overall} & \textbf{1} & \textbf{2} \\
\bottomrule\end{tabular}}
\end{minipage}\hfill\begin{minipage}[t]{0.56\linewidth}
\footnotesize\vspace{0pt}
\textbf{Lexical (vs.\ ground truth):}\\ ROUGE-L~0.33\quad BLEU-1~0.33\quad Semantic~0.56\par\smallskip
\textbf{Judge rationale:}\\ \textit{Premature ground frequency/taxi instruction before landing, invents frequency, nonstandard and unsafe; callsign correct.}
\end{minipage}
\end{tcolorbox}

\section{Significance Tests}
\label{app:stats}
Table~\ref{tab:stats} reports the hypothesis tests behind the claims in
Section~\ref{sec:results}. All are computed on per-model means across the nine models, so
each test treats the model as the unit of analysis. The in-context example and the
open-versus-closed gap are significant on similarity; ground-truth history is
significant only under the C5 prompt and on BLEU-1, consistent with its role as an
error-recovery rather than a per-turn effect.

\newcommand{\indentrow}[1]{\hspace{1.1em}#1}

\begin{table*}[!h]
\centering
\small
\setlength{\tabcolsep}{7pt}
\renewcommand{\arraystretch}{1.15}
\sisetup{
    detect-weight=true,
    detect-inline-weight=math
}

\caption{Hypothesis tests for the three experimental contrasts, computed on
per-model means across the nine models. Paired contrasts use an exact Wilcoxon
signed-rank test ($W$); the closed-versus-open contrast uses a one-sided
Mann--Whitney $U$ test. $\Delta$ is reported in the direction named in each
group heading, \emph{wins} counts models favoring that direction, and the last
column is the effect size ($d_z$ for the paired contrasts, Cohen's $d$ for the
unpaired one). Bold indicates $p<0.05$.}
\label{tab:stats}

\begin{tabular}{
    @{}l
    S[table-format=+1.3]
    c
    S[table-format=2.1]
    S[table-format=1.3]
    S[table-format=+1.2]
    @{}
}
\toprule
Measure & {$\Delta$} & {Wins} & {Statistic} & {$p$} & {Effect} \\
\midrule

\rowcolor{black!7}
\multicolumn{6}{@{}l@{}}{\itshape In-context example vs.\ none (paired, $W$, $n=9$)} \\

\indentrow{ROUGE-L (pooled)}         & +0.016 & 8/9 &  1.0 & {\bfseries 0.008} & +1.46 \\
\indentrow{Normalized BLEU (pooled)} & +0.009 & 8/9 &  1.0 & {\bfseries 0.008} & +1.47 \\
\indentrow{LLM-judge overall}        & +0.036 & 5/9 & 13.0 & 0.547             & +0.25 \\

\addlinespace[3pt]

\rowcolor{black!7}
\multicolumn{6}{@{}l@{}}{\itshape Ground-truth vs.\ self dialogue history (paired, $W$, $n=9$)} \\

\indentrow{ROUGE-L (pooled)}    & +0.010 & 5/9 & 14.0 & 0.359             & +0.49 \\
\indentrow{BLEU-1 (pooled)}     & +0.044 & 9/9 &  0.0 & {\bfseries 0.004} & +1.98 \\
\indentrow{ROUGE-L (prompt P1)} & -0.003 & 4/9 & 20.0 & 0.820             & -0.11 \\
\indentrow{ROUGE-L (prompt P5)} & +0.057 & 9/9 &  0.0 & {\bfseries 0.004} & +1.72 \\
\indentrow{LLM-judge overall}   & -0.078 & 3/9 & 14.5 & 0.371             & -0.40 \\

\addlinespace[3pt]

\rowcolor{black!7}
\multicolumn{6}{@{}l@{}}{\itshape Closed- vs.\ open-source (unpaired, $U$, 3 vs.\ 6)} \\

\indentrow{ROUGE-L} &
+0.059 &
\multicolumn{1}{c}{---} &
18.0 &
{\bfseries 0.012} &
+2.03 \\

\indentrow{LLM-judge overall} &
+1.050 &
\multicolumn{1}{c}{---} &
18.0 &
{\bfseries 0.012} &
+4.43 \\

\bottomrule
\end{tabular}
\end{table*}

\section{Per-Model Judge Scores}
\label{app:judge-table}
Table~\ref{tab:judge_per_model} gives every model's judge score on all seven dimensions,
averaged over conditions. The cross-dimension profile is uniform: callsign handling and
conciseness are high for all nine models while operational correctness is low, and the
open/closed separation is widest on correctness (2.71 versus 1.74) and hallucination
freedom (3.43 versus 2.24).

\begin{table*}[t]
\centering
\small
\caption{LLM-judge scores (1--5) per model, averaged over conditions and the top-25 judged outputs per cell. Best per column in bold.}
\label{tab:judge_per_model}
\begin{tabular}{lccccccc}
\toprule
Model & Phraseology & Correctness & Callsign & Conciseness & Safety & Halluc.-free & Overall \\
\midrule
\multicolumn{8}{l}{\textbf{Open-source}} \\
Qwen2.5-7B & 2.46 & 1.75 & 4.36 & 3.97 & 2.63 & 2.68 & 1.99 \\
Llama-3.1-8B & 2.40 & 1.74 & 3.65 & 3.88 & 2.51 & 2.22 & 1.89 \\
Gemma-2-9B & 2.80 & 1.94 & 4.14 & 4.34 & 2.80 & 2.37 & 2.12 \\
Qwen2.5-14B & 2.35 & 1.70 & \textbf{4.44} & 3.17 & 2.47 & 2.00 & 1.84 \\
Qwen2.5-32B & 2.81 & 2.00 & 4.29 & 3.83 & 2.83 & 2.33 & 2.20 \\
Mixtral-8x7B & 2.00 & 1.32 & 3.54 & 3.42 & 2.11 & 1.82 & 1.45 \\
\cmidrule(l){2-8}
\textit{Average} & \textit{2.47} & \textit{1.74} & \textit{4.07} & \textit{3.77} & \textit{2.56} & \textit{2.24} & \textit{1.91} \\
\addlinespace[3pt]
\multicolumn{8}{l}{\textbf{Closed-source}} \\
GPT-5.4 & \textbf{3.84} & \textbf{2.86} & 4.24 & \textbf{4.78} & \textbf{3.81} & \textbf{3.77} & \textbf{3.13} \\
GPT-5.4-mini & 3.59 & 2.57 & 4.15 & 4.61 & 3.54 & 3.20 & 2.82 \\
Claude-Sonnet-4.6 & 3.72 & 2.68 & 4.35 & 4.75 & 3.60 & 3.31 & 2.95 \\
\cmidrule(l){2-8}
\textit{Average} & \textit{3.71} & \textit{2.71} & \textit{4.25} & \textit{4.71} & \textit{3.65} & \textit{3.43} & \textit{2.96} \\
\bottomrule
\end{tabular}
\end{table*}

\section{Per-Model and Per-Prompt Breakdowns}
\label{app:breakdowns}
\begin{figure}[htbp]
  \centering
  \includegraphics[width=0.78\linewidth]{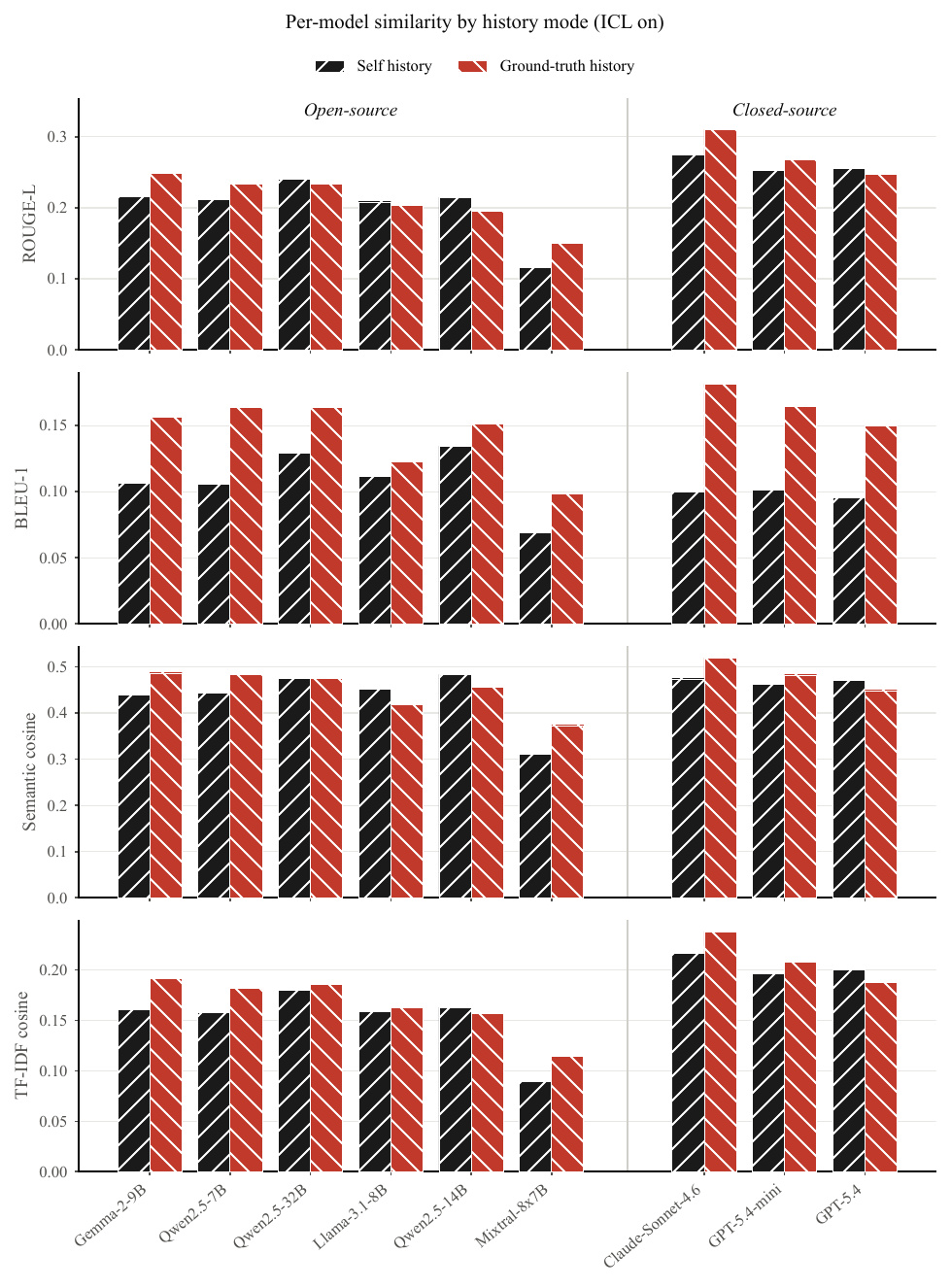}
  \caption{Per-model similarity by history mode (ICL on), grouped into open- and
  closed-source models. Bars are self history (hatched \texttt{//}) versus ground-truth
  history (hatched \texttt{\textbackslash\textbackslash}); panels are the four
  similarity metrics.}
  \label{fig:permodel_bars}
\end{figure}

\begin{figure}[htbp]
  \centering
  \includegraphics[width=0.78\linewidth]{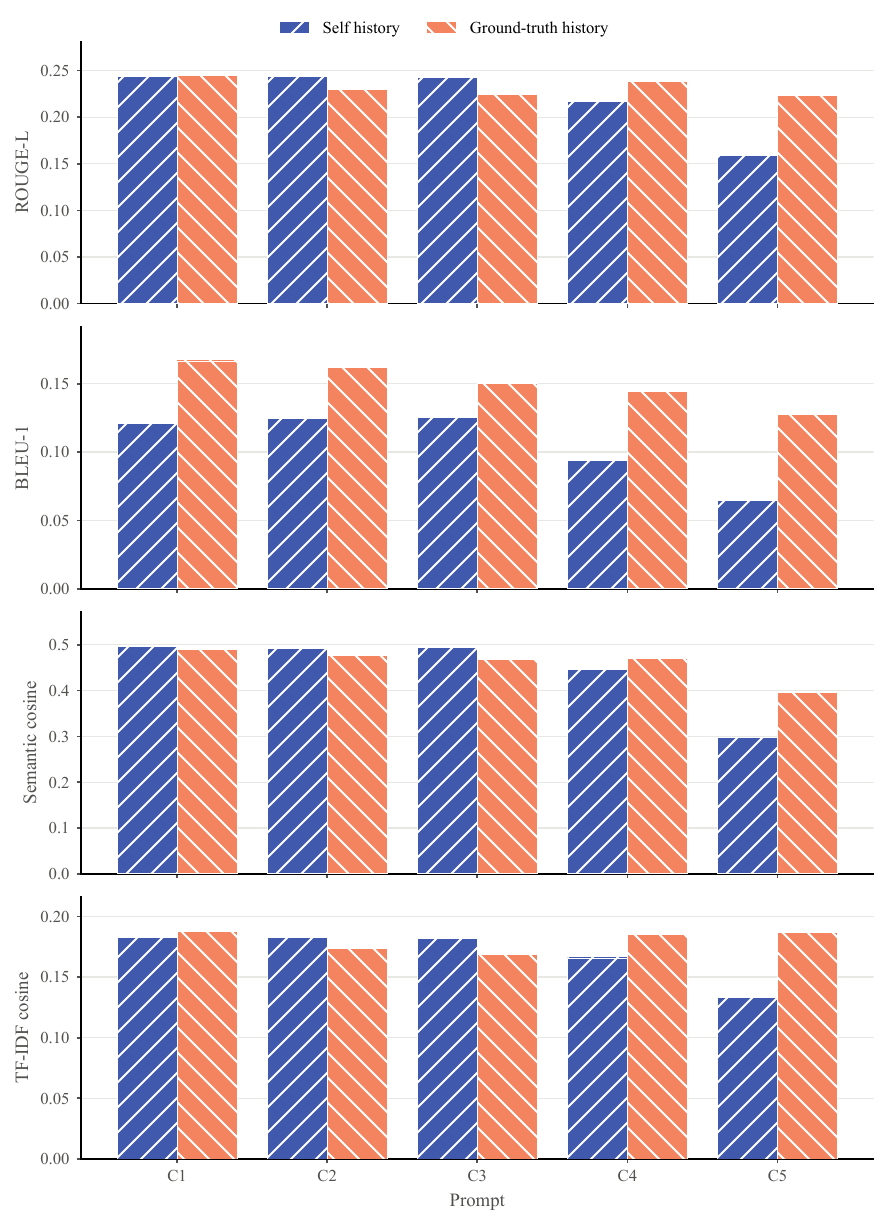}
  \caption{Per-prompt similarity by history mode (ICL on), across prompts C1--C5. Self
  history (indigo) collapses at C5 on every metric while ground-truth history (coral)
  remains flat, the per-prompt counterpart of Figure~\ref{fig:permodel_bars}.}
  \label{fig:perprompt_bars}
\end{figure}

Figures~\ref{fig:permodel_bars} and~\ref{fig:perprompt_bars} give the full breakdown
behind the pooled results of Section~\ref{sec:results}, both with in-context learning
enabled and bars contrasting self- against ground-truth history. Figure~\ref{fig:permodel_bars}
resolves the similarity by model, grouped into open- and closed-source; the
ground-truth-history advantage on BLEU-1 holds for every model, while on ROUGE-L it is
small and inconsistent. Figure~\ref{fig:perprompt_bars} resolves the same quantities by
prompt, and isolates the C5 collapse: self-history (indigo) falls sharply at C5 on all
four metrics while ground-truth history (coral) holds, the per-prompt view of the
error-accumulation effect discussed in Section~\ref{sec:res-grounding}.

\clearpage
\section{Agreement Analysis}
\label{app:agreement}
\begin{figure}[!t]
  \centering
  \begin{subfigure}[t]{0.49\linewidth}
    \centering
    \includegraphics[width=\linewidth]{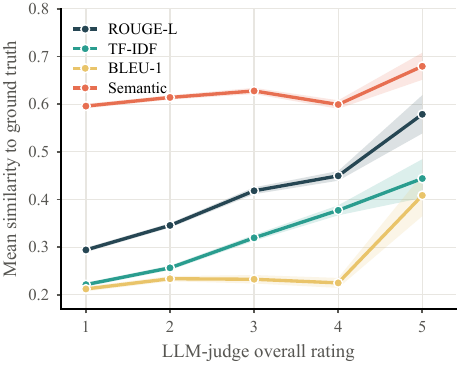}
    \caption{Similarity vs.\ judge rating.}
    \label{fig:sim_by_rating}
  \end{subfigure}\hfill
  \begin{subfigure}[t]{0.49\linewidth}
    \centering
    \includegraphics[width=\linewidth]{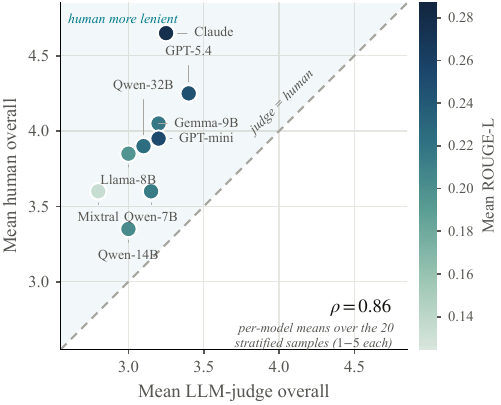}
    \caption{Judge vs.\ human, per model.}
    \label{fig:jh_models}
  \end{subfigure}
  \caption{Agreement between the evaluation layers. (a)~Mean automatic similarity as a
  function of the LLM-judge overall rating (sample level, $n=4{,}500$): ROUGE-L and
  TF-IDF rise monotonically with judge rating, BLEU-1 responds only at rating~5, and
  embedding similarity is nearly flat. (b)~Per-model mean judge rating against mean
  human rating on the 180 annotated outputs (color = mean ROUGE-L); every model lies
  above the identity line, so the human is the more lenient rater for all nine.}
  \label{fig:agreement_pair}
\end{figure}

\begin{figure}[!t]
  \centering
  \begin{subfigure}[t]{0.9\linewidth}
    \centering
    \includegraphics[width=\linewidth]{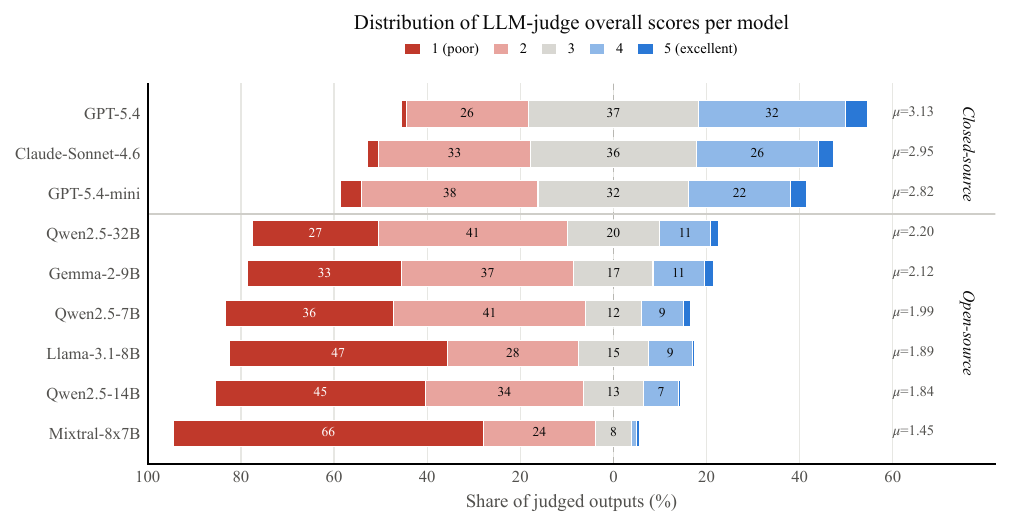}
    \caption{LLM judge (full judged subset, $n=4{,}500$).}
    \label{fig:sd_judge}
  \end{subfigure}

  \vspace{6pt}

  \begin{subfigure}[t]{0.9\linewidth}
    \centering
    \includegraphics[width=\linewidth]{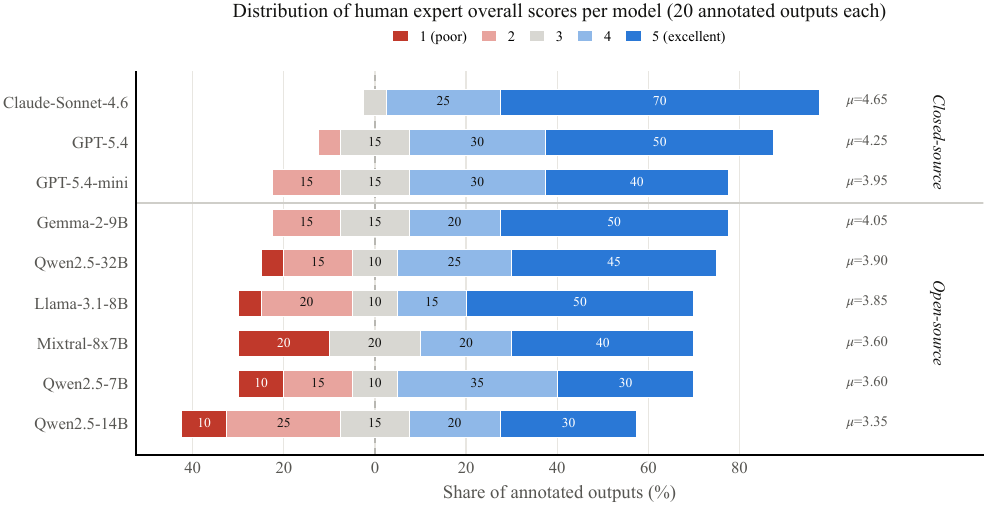}
    \caption{Human expert (180 annotated outputs, 20 per model).}
    \label{fig:sd_human}
  \end{subfigure}
  \caption{Distribution of overall scores per model, as diverging Likert bars.
  Relative to the judge (a), the human distribution (b) shifts markedly toward 4 and 5,
  the same leniency offset quantified in Table~\ref{tab:agreement_dims}. Note the
  differing sample sizes: the judge panel is over the full judged subset, the human
  panel over the 180-sample validation set.}
  \label{fig:score_dists}
\end{figure}

This appendix collects the metric--judge and judge--human agreement evidence summarised
in Sections~\ref{sec:res-metrics} and~\ref{sec:res-human}. Table~\ref{tab:corr_matrix}
is the rank-correlation matrix over all six measures; Table~\ref{tab:agreement} reports
every layer-pair correlation at sample and model level; Table~\ref{tab:agreement_dims}
breaks the judge--human comparison down by rubric dimension and by judge band. The
figures show how similarity rises with judge rating at the sample level
(Fig.~\ref{fig:sim_by_rating}), the judge--human means per model
(Fig.~\ref{fig:jh_models}), and the two raters' overall-score distributions
(Fig.~\ref{fig:score_dists}). Two patterns recur: the reference-overlap metrics
(ROUGE-L, TF-IDF) are the only automatic measures that track either human-facing rater,
and the human sits consistently above the judge, most on the outputs the judge rates
lowest.


\begin{table}[t]
\centering
\small
\caption{Spearman rank-correlation matrix over all six evaluation measures, computed on the 180 human-annotated outputs so that every measure is available for every observation. The two reference-overlap metrics (ROUGE-L, TF-IDF) are the only automatic measures that track either human-facing judgement.}
\label{tab:corr_matrix}
\begin{tabular}{lccccc}
\toprule
 & ROUGE-L & TF-IDF & BLEU-1 & Semantic & LLM judge \\
\midrule
TF-IDF & 0.73 &  &  &  &  \\
BLEU-1 & 0.58 & 0.36 &  &  &  \\
Semantic & 0.32 & 0.47 & 0.30 &  &  \\
LLM judge & 0.53 & 0.49 & 0.02 & 0.10 &  \\
Human & 0.34 & 0.38 & -0.05 & 0.09 & 0.67 \\
\bottomrule
\end{tabular}
\end{table}

\begin{table}[t]
\centering
\small
\caption{Agreement between the three evaluation layers. Sample-level statistics treat each scored output as one observation; model-level statistics use per-model means ($n{=}9$). $\kappa_w$ is quadratic-weighted Cohen's kappa, defined only when both measures share the 1--5 ordinal scale.}
\label{tab:agreement}
\begin{tabular}{llccc}
\toprule
Comparison & Level & Pearson $r$ & Spearman $\rho$ & $\kappa_w$ \\
\midrule
\multicolumn{5}{l}{\textbf{Automatic metric vs.\ LLM-judge overall} (sample $n{=}4{,}500$)} \\
ROUGE-L & sample & 0.51 & 0.52 & -- \\
ROUGE-L & model & 0.90 & 0.93 & -- \\
TF-IDF & sample & 0.50 & 0.48 & -- \\
TF-IDF & model & 0.91 & 0.90 & -- \\
BLEU-1 & sample & 0.12 & 0.06 & -- \\
BLEU-1 & model & 0.43 & 0.22 & -- \\
Semantic & sample & 0.08 & 0.09 & -- \\
Semantic & model & 0.67 & 0.50 & -- \\
\addlinespace[3pt]
\multicolumn{5}{l}{\textbf{Automatic metric vs.\ human overall} (sample $n{=}180$)} \\
ROUGE-L & sample & 0.31 & 0.34 & -- \\
ROUGE-L & model & 0.73 & 0.81 & -- \\
TF-IDF & sample & 0.32 & 0.38 & -- \\
TF-IDF & model & 0.72 & 0.79 & -- \\
BLEU-1 & sample & 0.02 & -0.05 & -- \\
BLEU-1 & model & 0.17 & -0.03 & -- \\
Semantic & sample & 0.09 & 0.09 & -- \\
Semantic & model & 0.39 & 0.28 & -- \\
\addlinespace[3pt]
\multicolumn{5}{l}{\textbf{LLM judge vs.\ human annotator} (sample $n{=}180$)} \\
Overall rating & sample & 0.71 & 0.67 & 0.60 \\
Overall rating & model & 0.69 & 0.86 & -- \\
\bottomrule
\end{tabular}
\end{table}

\begin{table}[t]
\centering
\small
\caption{Per-dimension and per-stratum agreement between the LLM judge and the human annotator ($n{=}180$). $\Delta$ is the human mean minus the judge mean; positive values mean the human is more lenient. Within-stratum correlations are attenuated by range restriction and $\kappa_w$ is uninformative there, so it is reported only for the pooled sample.}
\label{tab:agreement_dims}
\begin{tabular}{lccccc}
\toprule
Dimension & Judge mean & Human mean & $\Delta$ & Spearman $\rho$ & $\kappa_w$ \\
\midrule
Phraseology & 3.24 & 4.30 & +1.06 & 0.45 & 0.36 \\
Correctness & 2.89 & 3.79 & +0.90 & 0.63 & 0.52 \\
Callsign & 4.43 & 4.86 & +0.43 & 0.35 & 0.43 \\
Conciseness & 4.34 & 4.51 & +0.17 & 0.41 & 0.59 \\
Safety & 3.63 & 3.91 & +0.28 & 0.53 & 0.59 \\
Halluc.-free & 3.61 & 3.96 & +0.34 & 0.48 & 0.46 \\
\textbf{Overall} & 3.12 & 3.91 & +0.79 & 0.67 & 0.60 \\
\midrule
\multicolumn{6}{l}{\emph{Overall rating, split by judge stratum}} \\
Low ($n$=45) & 1.13 & 2.58 & +1.44 & 0.44 & -- \\
Mid ($n$=45) & 2.87 & 3.78 & +0.91 & 0.36 & -- \\
High ($n$=90) & 4.24 & 4.64 & +0.40 & 0.20 & -- \\
\bottomrule
\end{tabular}
\end{table}


\clearpage
\section{Latency and Run-to-Run Variability}
\label{app:extras}
\begin{figure}[h]
  \centering
  \begin{subfigure}[t]{0.49\linewidth}
    \centering
    \includegraphics[width=\linewidth]{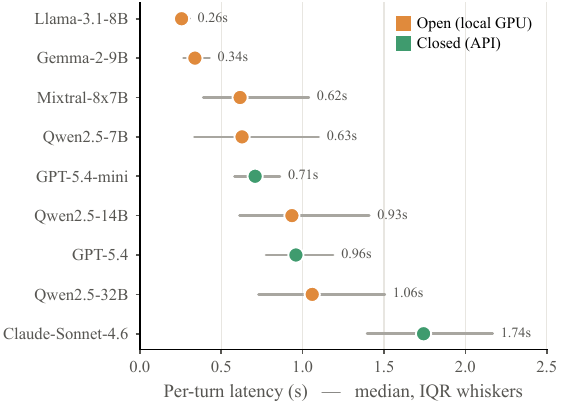}
    \caption{Median per-turn latency.}
    \label{fig:latency}
  \end{subfigure}\hfill
  \begin{subfigure}[t]{0.49\linewidth}
    \centering
    \includegraphics[width=\linewidth]{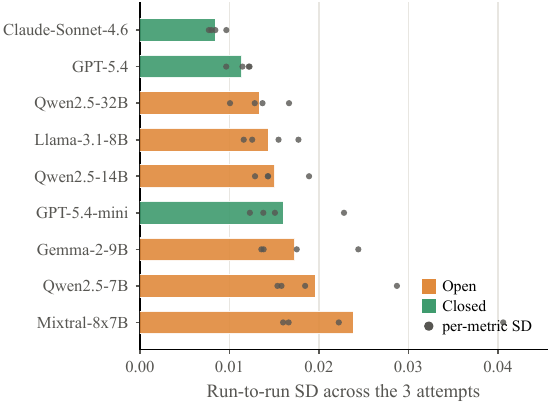}
    \caption{Run-to-run variability.}
    \label{fig:variability}
  \end{subfigure}
  \caption{Serving cost and stability per model. Open models run locally on a GPU;
  closed models are API calls. (a)~Median per-turn latency with interquartile-range
  whiskers, spanning 0.26\,s (Llama-3.1-8B) to 1.74\,s (Claude-Sonnet-4.6).
  (b)~Standard deviation across the three attempts; bars are the mean over the four
  metrics, dots the individual per-metric values ($\leq 0.04$ throughout).}
  \label{fig:serving}
\end{figure}

Figure~\ref{fig:latency} reports median per-turn latency and
Figure~\ref{fig:variability} the standard deviation across the three attempts. Latency
spans 0.26\,s (Llama-3.1-8B on a local GPU) to 1.74\,s (Claude-Sonnet-4.6 via API);
variability is small for every model ($\leq 0.04$ on all four metrics), confirming that
the rankings in the main text are stable across repetitions.

\end{document}